\documentclass[11pt,a4paper]{article}

\usepackage[T1]{fontenc}
\usepackage{lmodern}
\usepackage[margin=1in]{geometry}

\usepackage{amsmath,amssymb,amsthm}
\usepackage{ifpdf}
\usepackage{algorithm}

\usepackage{lipsum}
\usepackage{amsfonts}
\usepackage{graphicx}
\usepackage[caption=false]{subfig} 
\usepackage{algorithmic}

\usepackage[hidelinks]{hyperref}
\usepackage[capitalize,nameinlink]{cleveref}

\theoremstyle{plain}
\newtheorem{theorem}{Theorem}[section]
\newtheorem{lemma}[theorem]{Lemma}

\theoremstyle{definition}

\newtheorem{assumption}[theorem]{Assumption}

\theoremstyle{remark}

\crefname{prop}{Proposition}{Propositions}
\crefname{coro}{Corollary}{Corollaries}
\crefname{hypothesis}{Hypothesis}{Hypotheses}
\crefname{fact}{Fact}{Facts}
\crefname{assumption}{Assumption}{Assumptions}

\AddToHook{cmd/appendix/before}{%
  \crefalias{section}{appendix}%
  \crefalias{subsection}{appendix}%
}

\title{Adaptive subspace modeling with functional Tucker}

\title{Adaptive subspace modeling\\ with functional Tucker decomposition}

\newcommand{\email}[1]{\href{mailto:#1}{\texttt{#1}}}
\date{}

\author{Noah Steidle\thanks{KU Leuven 
    (\email{noahmaximilian.steidle@kuleuven.be}).}
\and Joppe De Jonghe\thanks{KU Leuven 
    (\email{joppe.dejonghe@kuleuven.be}).}
\and Mariya Ishteva\thanks{KU Leuven
    (\email{mariya.ishteva@kuleuven.be}).}
    }
\usepackage{amsopn}

\usepackage{color} 
\usepackage{soul} 

\usepackage{booktabs}
\usepackage{multirow}
\usepackage{makecell}
\usepackage{threeparttable}
\usepackage{xcolor}
\usepackage{colortbl}
\usepackage{bm}
\usepackage{enumerate}
\usepackage{enumitem}

\DeclareMathOperator{\subjto}{subj. to }
\DeclareMathOperator{\vecc}{vec}
\DeclareMathOperator{\linspan}{span}
\DeclareMathAlphabet\mathbfcal{OMS}{cmsy}{b}{n}
\DeclareMathOperator{\Tr}{Tr}
\NewDocumentCommand{\trans}{m}{#1^{\mathsf{T}}}

\definecolor{rowgray}{gray}{0.93}

\ifpdf
\hypersetup{
  pdftitle={Adaptive subspace modeling with functional Tucker decomposition},
  pdfauthor={N. Steidle, J. De Jonghe, and M. Ishteva}
}
\fi

\begin{document}

\maketitle


\begin{abstract}
Tensors provide a structured representation for multidimensional data, yet discretization discards the underlying continuous structure when the data originate from continuous processes. We address this limitation by introducing a functional Tucker decomposition (FTD) that embeds a mode-wise continuity constraint directly into the factorization. The FTD models the continuous mode as a function in a reproducing kernel Hilbert space (RKHS), avoiding a prespecified basis while preserving the multilinear subspace structure of the Tucker model. We derive a reconstruction error bound for the continuous mode that quantifies the approximation quality when a subspace estimated on one domain is reused on another. This bound provides theoretical justification for subspace transfer, whose practical value we demonstrate on cross-domain classification tasks in hyperspectral imaging and multivariate time-series analysis.
\end{abstract}



\section{Introduction}\label{sec:introduction}

Tensors and tensor decompositions have been used in a wide variety of applications to model high-dimensional discrete data or measurements, structured in a tabular scheme. Examples of tensor decompositions include the canonical polyadic decomposition (CPD) and Tucker decomposition (TD), which have been widely applied in signal processing, chemometrics and numerical analysis contexts \cite{BallardKolda2025}. 
However, in practice, discrete data captured in the tensor may originate from a continuous system which is better represented in a tensor with one or more continuous modes. Mathematically, this structure is called a \textit{quasitensor}, generalizing the notion of \textit{quasimatrices} \cite{stewart1998}.

Standard tensor decompositions such as CPD and TD do not take into account the continuity of quasitensors originating from smooth data processes, effectively losing this information. To this end, several publications have aimed at addressing the idea of functional modes in the context of the CPD and TD. For instance, the authors of \cite{timmerman2002three} proposed to constrain the elements of the functional factor matrix to come from a smooth B-spline basis. Beylkin et al. \cite{beylkin2005algorithms} introduced a CPD model where all the modes are functional which corresponds to functional regression in three dimensions and is related to the work of Dolgov et al. \cite{dolgov2021functional} who apply the same idea to the TD with Chebyshev interpolation for the functional modes. Also, structured data fusion proposed in \cite{SorberVanBarelDeLathauwer2015} provides a general optimization and modeling framework which includes mode-wise smoothness constraints via its factor-structure transformations. More recent examples include the work of Tang, Larsen, Han and Kolda ~\cite{TangKoldaZhang2026,LarsenEtAl2024,HanShiZhang2024} which culminated in the CP-HiFi decomposition that allows arbitrary numbers of continuous modes with a kernel-based functional representation. Moreover, Qian et al. \cite{QianEtAl2025} approached a smooth Tucker tensor decomposition by adding temporal smoothing penalties to better model continuous measurements of blood pressure and heart rates.

\begin{figure}[tbhp]
    \centering
    \includegraphics[width=0.95\linewidth]{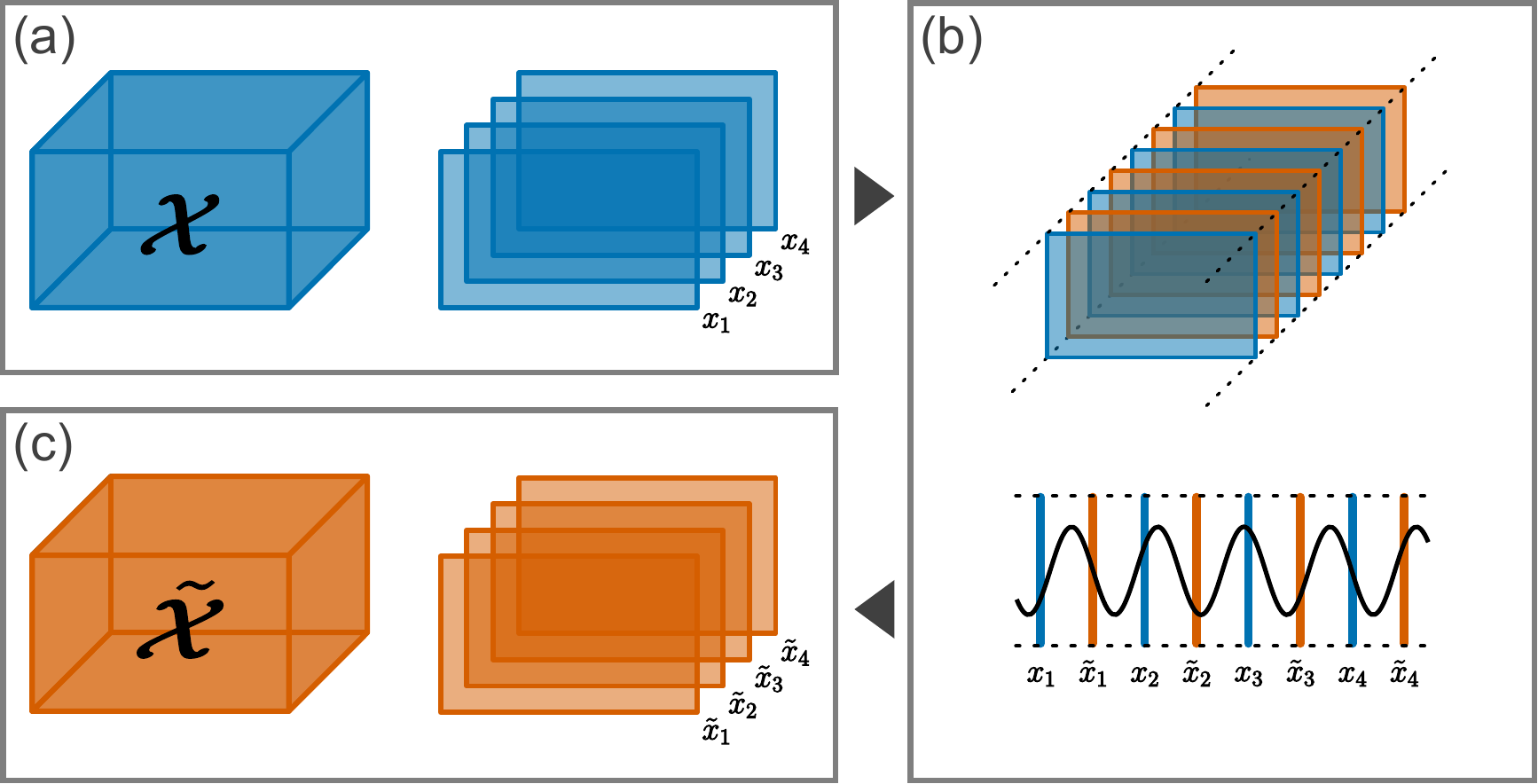}
    \caption{Overview of the adaptive subspace modeling framework via functional resampling. (a) An input tensor $\mathbfcal{X}$ is interpreted as slices in the third mode corresponding to sampling points $x_1,\ldots,x_4$. (b) The mode-3 fibers are represented as continuous functions of the sampling coordinates, allowing evaluation at arbitrary points, here new points $\widetilde{x}_1,\ldots,\widetilde{x}_4$. (c) Evaluating the continuous representation at the new sampling points yields a new tensor $\widetilde{\mathbfcal{X}}$, modeling $\mathbfcal{X}$ on a target domain in the functional mode. In practice, the construction and re-construction of the tensor is realized via the functional Tucker decomposition (FTD).}
    \label{fig:cube_subspace}
\end{figure}

In this work we investigate a functional Tucker decomposition (FTD), driven by the theoretical foundation of quasitensors and RKHS. The aim of this decomposition is to extract subspace information in each mode while retaining continuity in a specified mode. By this setup, the FTD enables adaptive subspace modeling, i.e., the subspace transfer from the input domain to a specified target domain. Leveraging the conservation of subspace information over the transfer, this method proves beneficial in the context of tensor cross-domain classification. In particular, the functional representation allows to construct a tailored classification model at test time, incorporating \textit{a priori} knowledge about the functional mode, compared to a fixed model used during testing. The framework of adaptive subspace modeling is illustrated in \Cref{fig:cube_subspace}.
While functional tensor decompositions have been considered in the literature, our method complements recent advances by building on the Tucker decomposition which -- unlike the CP decomposition -- permits mode-wise rank selection and represents inter-modal interactions through a dense core tensor. Moreover, while \cite{dolgov2021functional} primarily focuses on efficient storage of approximations of functions with the Tucker decomposition, our method instead emphasizes the flexible and adaptive representation in the RKHS framework, which facilitates subspace transfer.

In the following, we outline notation and technical preliminaries such as the methodology of TD-based classification and RKHS theory in \Cref{sec:notation_preliminaries}. The model and optimization formulation for the functional Tucker decomposition, an algorithm to solve the optimization problem, as well as an error bound on the reconstruction error in the transfer domain are presented in \Cref{sec:functional_tucker_decomposition}. \Cref{sec:experiments} provides experimental results with semi-synthetic data and two real-world datasets that showcase the benefits of combining Tucker decomposition with functional adaptability in the context of classification. Finally, conclusions and future work are discussed. 

\section{Notation and preliminaries}\label{sec:notation_preliminaries}
\subsection{Tensors and Tucker decomposition}\label{subsec:tensors_tucker}

In the following, vectors are denoted by bold lowercase Roman letters (e.g., $\textbf{a},\textbf{b}$), matrices by bold uppercase Roman letters (e.g., $\textbf{A},\textbf{B}$) and tensors by bold uppercase calligraphy letters (e.g., $\mathbfcal{X},\mathbfcal{Y}$). Following \cite{BallardKolda2025}, a tensor $\mathbfcal{X}\in\mathbb{R}^{m \times n \times p}$ is said to be of \textit{order} three, referring to its number of dimensions, also known as \textit{modes}. The \textit{slices} of $\mathbfcal{X}$ are its $2$-way subtensors which can be obtained by fixing all but two indices of the tensor. For $k\in[3]:=\{1,2,3\}$, the mode-$k$ \textit{fibers} of $\mathbfcal{X}$ are defined by the $1$-way subtensors obtained by fixing every but the $k$th mode index. $\textbf{X}_{(k)}$ denotes the unfolding of $\mathbfcal{X}$ over its $k$th mode, as described in \cite{BallardKolda2025}, i.e., the columns of a mode-$k$ unfolding are the mode-$k$ tensor fibers. The \textit{tensor-times-matrix product} or \textit{$k$-mode product} $\mathbfcal{Y} = \mathbfcal{X} \bullet_k \textbf{A}$ defines a mode-wise multiplication of a tensor $\mathbfcal{X}$ with a matrix $\mathbfcal{A}$ via $\textbf{Y}_{(k)} = \textbf{A}\textbf{X}_{(k)}$.

Let $||\cdot||$ denote the norm which equals the square root of the sum of all squared entries. For a tensor $\mathbfcal{X}$, it holds $||\mathbfcal{X}|| = \sqrt{\sum_{i=1}^{m}\sum_{j=1}^{n}\sum_{k=1}^{p} x_{ijk}^2}$. Given a positive definite matrix $\textbf{K}\in\mathbb{R}^{p \times p}$, i.e., $\trans{\textbf{x}}\textbf{K}\textbf{x} > 0$ for all $\textbf{x}\neq0$, define the \textit{weighted matrix norm} via
\begin{align}
    ||\textbf{W}||_{\textbf{K}}^{2} = \sum_{i=1}^{s} \textbf{w}_{i}^{\mathsf{T}} \textbf{K} \textbf{w}_i = 
    \Tr(\trans{\textbf{W}}\textbf{K}\textbf{W})
    \label{eq:def_weighted_matrix_norm}
\end{align}
for a matrix $\textbf{W}\in\mathbb{R}^{p \times s}$. Here, $\Tr(\cdot)$ denotes the trace operator that computes the sum of the elements on the main diagonal of a matrix. Finally, $\otimes$ denotes the Kronecker product. According to \cite{BallardKolda2025}, the following properties hold:
\begin{align}
    \trans{\left( \textbf{A} \otimes \textbf{B} \right)} &= \trans{\textbf{A}} \otimes \trans{\textbf{B}},\label{eq:kron_transpose}\\
    \left( \textbf{A} \otimes \textbf{B} \right) \left( \textbf{C} \otimes \textbf{D} \right) &= \left( \textbf{A}\textbf{C} \right) \otimes \left( \textbf{B}\textbf{D} \right),\label{eq:kron_ABCD}\\
    \vecc\left(\textbf{A}\textbf{C}\trans{\textbf{B}}\right) &= (\textbf{B} \otimes \textbf{A}) \vecc(\textbf{C}).\label{eq:mixed_kron_matr_vec_product}
\end{align}

Classical \textit{tabular tensors} represent discrete measurements within their modes. Lifting this restriction to allow continuous processes in part of the modes yields the concept of \textit{quasitensors} which naturally generalize the notion of \textit{quasimatrices} \cite{stewart1998}. A quasimatrix $\widetilde{\textbf{A}} \in \mathbb{R}^{\infty \times s}$ describes an ordered set of $s$ univariate functions,
\begin{align*}
    \widetilde{\textbf{A}} = [ a_1 \cdots a_s ] \text{ where } a_\gamma : \mathbb{R} \rightarrow \mathbb{R} \text{ for } \gamma \in [s].
\end{align*}
Evaluating the quasimatrix $\widetilde{\textbf{A}}$ on a set of $p$ evaluation points $I = \{x_1, \ldots, x_p\} \subset \mathbb{R}$ yields a (tabular) matrix $\widehat{\textbf{A}}_{I}\in\mathbb{R}^{p \times s}$ as follows
\begin{align*}
    \widehat{\textbf{A}}_{I} = 
    \begin{bmatrix}
        a_1(x_1) & a_2(x_1) & \cdots & a_s(x_1) \\
        a_1(x_2) & a_2(x_2) & \cdots & a_s(x_2) \\
        \vdots   & \vdots   & \ddots & \vdots   \\
        a_1(x_p) & a_2(x_p) & \cdots & a_s(x_p)
    \end{bmatrix}\in\mathbb{R}^{p \times s}.
\end{align*}
An analogous approach expands the concept of tensors to quasitensors $\widetilde{\mathbfcal{T}}\in\mathbb{R}^{m \times n \times \infty}$ and their evaluated counterpart $\widehat{\mathbfcal{T}}_{I}\in\mathbb{R}^{m \times n \times p}$. In the following, we consider quasitensors that consist of two discrete modes and one continuous mode. Nonetheless, the number of discrete and continuous modes can be expanded arbitrarily. Without loss of generality, this work considers the continuous mode of a tensor to be the last one.

Similar to matrix decompositions such as the singular value decomposition (SVD), tensors can be decomposed. The \textit{rank-$(q,r,s)$ Tucker decomposition} deconstructs a tensor $\mathbfcal{X}\in\mathbb{R}^{m \times n \times p}$ into a \textit{core tensor} $\mathbfcal{G}\in\mathbb{R}^{q \times r \times s}$ and \textit{factor matrices} $\textbf{A}\in\mathbb{R}^{m \times q}$, $\textbf{B}\in\mathbb{R}^{n \times r}$ and $\textbf{C}\in\mathbb{R}^{p \times s}$ such that $\mathbfcal{X} \approx \mathbfcal{G} \bullet_1 \textbf{A} \bullet_2 \textbf{B} \bullet_3 \textbf{C} \label{eq:def_tucker}$ \cite{Hitchcock1927,Tucker1966}. An alternative way to express the Tucker decomposition is $\mathbfcal{X} \approx [\![\mathbfcal{G};\textbf{A},\textbf{B},\textbf{C}]\!]$. Given an exact Tucker decomposition, i.e., $\mathbfcal{X}=[\![ \mathbfcal{G}; \textbf{A}, \textbf{B}, \textbf{C}]\!]$, its unfoldings can be directly computed via
\begin{align}
    &\textbf{X}_{(1)} 
    = \textbf{A}\mathbf{G}_{(1)}\trans{\left( \textbf{C} \otimes \textbf{B} \right)}
    = \textbf{A} \left( \mathbfcal{G} \bullet_2 \textbf{B} \bullet_3 \textbf{C} \right)_{(1)}\label{eq:tucker_unfolding1}\\
    &\textbf{X}_{(2)} 
    = \textbf{B}\mathbf{G}_{(2)}\trans{\left( \textbf{C} \otimes \textbf{A} \right)}
    = \textbf{B} \left( \mathbfcal{G} \bullet_1 \textbf{A} \bullet_3 \textbf{C} \right)_{(2)}\label{eq:tucker_unfolding2}\\
    &\textbf{X}_{(3)} 
    = \textbf{C}\mathbf{G}_{(3)}\trans{\left( \textbf{B} \otimes \textbf{A} \right)}
    = \textbf{C} \left( \mathbfcal{G} \bullet_1 \textbf{A} \bullet_2 \textbf{B} \right)_{(3)}\label{eq:tucker_unfolding3}.
\end{align}

In general, solving for the Tucker decomposition jointly for all factor matrices is challenging \cite{BallardKolda2025}. Consequently, optimization formulations typically rely on alternating least-squares (ALS) methods that estimate one factor matrix at a time through mode‑wise optimization, followed by the computation of the core tensor in a final step. Algorithms based on this framework include the higher‑order singular value decomposition (HOSVD), the sequentially truncated HOSVD (ST‑HOSVD), and higher‑order orthogonal iteration (HOOI) \cite{BallardKolda2025}.

\subsection{Classification using HOSVD}\label{subsec:class_hosvd}

While the subspace transfer method proposed in this work is independent of any particular use case, the experiments in \Cref{sec:experiments} demonstrate its effectiveness through a series of classification tasks. To exploit the tensor structure of the data, classification is performed via a multidimensional variant of \textit{Soft Independent Modeling by Class Analogy} (SIMCA) \cite{Wold1976}, based on the formulation introduced by Savas and Eldén \cite{SavasElden2007}. 

The method proceeds in two stages, both grounded in the Tucker decomposition. In the training stage, for each class $\mu$, the tensor $\mathbfcal{X}^{\mu} \in \mathbb{R}^{M_\mu \times n \times p}$ of stacked training samples is decomposed via a truncated HOSVD, yielding an orthonormal set of $k$ dictionary matrices $\{\mathbf{D}_\nu^\mu\}_{\nu=1}^k$ spanning a dominant low-dimensional subspace characteristic of the class. In the test stage, an unlabeled sample $\mathbf{Y} \in \mathbb{R}^{n \times p}$ is approximated within each class model. Thanks to the HOSVD, the Tucker core is orthogonal and the residual simplifies to $R(\mu) = 1 - \sum_{\nu=1}^k \langle \mathbf{Y}, \mathbf{D}_\nu^\mu \rangle^2.$
The sample is then assigned to the class $\mu^*$ for which $R(\mu^*)$ is minimal. A more detailed description to the method can be found in \Cref{app:subsec:class_hosvd}.

\subsection{Reproducing kernel Hilbert spaces (RKHS)}\label{subsec:rkhs}

Given a set of points $I = \{x_1, \ldots, x_p\} \subset \mathbb{R}$ and a symmetric positive semidefinite kernel $\mathcal{K}: I \times I \rightarrow \mathbb{R}$, the RKHS is defined as
\begin{align*}
    \mathcal{H}_\mathcal{K} := \overline{\linspan\{\mathcal{K}(\cdot, x_i) \mid x_i \in I\}},
\end{align*}
equipped with the inner product $\langle f, g \rangle_\mathcal{K} = \sum_{i,j=1}^{p} \alpha_i \beta_j \mathcal{K}(x_i, x_j)$ for functions $f = \sum_i \alpha_i \mathcal{K}(\cdot, x_i)$ and $g = \sum_j \beta_j \mathcal{K}(\cdot, x_j)$ \cite{SchoelkopfSmola2001}. There are many different choices of kernels which generate a RKHS. The kernel used in this work is the standard \textit{Gaussian kernel} which is defined via $\mathcal{K}(x_i,x_j) = \exp\left(\frac{-(x_i-x_j)^2}{2d^2}\right)$ for a parameter $d\neq0$. The Gaussian kernel is in fact positive definite \cite{Wendland2004}. Since positive definiteness implies positive semidefiniteness, the Gaussian is admissible in the sense above and induces the RKHS $\mathcal{H}_\mathcal{K}$.

The key property of these spaces is the \textit{reproducing property} $\langle \mathcal{K}(\cdot, x_i), f \rangle_\mathcal{K} = f(x_i)$, enabling pointwise evaluation via inner products \cite{SchoelkopfSmola2001}. Furthermore, the Representer theorem \cite{KimeldorfWahba1971, SchoelkopfSmola2001} guarantees that the minimizer of the regularized least-squares problem
\begin{align}
    \min_{f\in\mathcal{H}_{\mathcal{K}}} \frac{1}{p}\sum_{i=1}^{p}(y_i - f(x_i))^2 + \frac{\lambda}{2}\|f\|_\mathcal{K}^2, \quad \lambda > 0,\label{eq:min_req_ls}
\end{align}
admits a finite-dimensional representation
\begin{align}
    f^*(\cdot) = \sum_{i=1}^{p} w_i \mathcal{K}(\cdot, x_i) = \widetilde{\mathbf{K}}\mathbf{w},\label{eq:min_req_ls_sol}
\end{align}
where $\mathbf{w} \in \mathbb{R}^p$ and $\widetilde{\mathbf{K}} = \mathcal{K}(\cdot, \mathbf{x}) \in \mathbb{R}^{\infty \times p}$ is a quasimatrix that reduces to the kernel matrix $\mathbf{K} \in \mathbb{R}^{p \times p}$ when evaluated at the design points in $I$. This representation is central to the optimization problem in \Cref{sec:functional_tucker_decomposition}. As before, more information can be found in \Cref{app:subsec:rkhs}.

\section{Functional Tucker decomposition}\label{sec:functional_tucker_decomposition}

In this section we derive a Functional Tucker decomposition (FTD) by employing the theory of reproducing kernel Hilbert spaces (RKHS). First, we incorporate the RKHS formulation into the Tucker model by adjusting the corresponding optimization problem. Second, we derive closed form solutions for each variable of the optimization problem individually. The solutions are then merged into an algorithm. Finally, we provide a theoretical argument for the subspace transfer with the reconstruction error bound for the continuous mode that quantifies the approximation quality within a shift of subspaces.

The following derivation is based on input tensors with two discrete modes and one continuous mode. Nonetheless, the theory can be generalized to arbitrary numbers of discrete and continuous modes thanks to the mode-wise perspective of the optimization problem.

\subsection{Setup}\label{subsec:setup}
Let $m,n,q,r,s \in\mathbb{N}$ with $q \leq m$ and $r \leq n$ and $I\subset\mathbb{R} $. Finding a functional rank-($q$,$r$,$s$) Tucker decomposition, translates to decomposing a quasitensor $\widetilde{\mathbfcal{T}}\in\mathbb{R}^{m \times n \times \infty}$ as
\begin{align}
    \widetilde{\mathbfcal{T}} = [\![ \mathbfcal{G}; \textbf{A},\textbf{B},\widetilde{\textbf{C}} ]\!]\label{eq:ftd_element_quasi}
\end{align}
for a core tensor $\mathbfcal{G}\in\mathbb{R}^{q \times r \times s}$, factor matrices $\textbf{A} = [\textbf{a}_{1} \cdots \textbf{a}_q]\in\mathbb{R}^{m \times q}$ and $\textbf{B} = [\textbf{b}_{1} \cdots \textbf{b}_{r}]\in \mathbb{R}^{n \times r}$, as well as a quasimatrix $\widetilde{\textbf{C}}\in \mathbb{R}^{\infty \times s}$. 


In practice, a tensor like above is considered for a discrete set of measurements in the continuous mode which yields a discrete tabular tensor. Therefore, define the tensor $\widehat{\mathbfcal{T}}_{I}\in\mathbb{R}^{m \times n \times p}$ as the quasitensor $\widetilde{\mathbfcal{T}}$ evaluated at a set of $p$ design points $I = \{x_1, \ldots,x_p\} \subset \mathbb{R}$ in the third mode. This yields a classical rank-($q$,$r$,$s$) Tucker decomposition by evaluating \Cref{eq:ftd_element_quasi} according to
\begin{align*}
    \widehat{\mathbfcal{T}}_{I} = [\![ \mathbfcal{G}; \textbf{A},\textbf{B},\widehat{\textbf{C}}_{I} ]\!],
\end{align*}
where $\widehat{\textbf{C}}_{I}\in\mathbb{R}^{p \times s}$ is the quasimatrix evaluated at the design points in $I$, i.e., 
\begin{align*}
    \widehat{\textbf{C}}_{I} = 
    \begin{bmatrix}
        c_1(x_1) & c_2(x_1) & \cdots & c_s(x_1) \\
        c_1(x_2) & c_2(x_2) & \cdots & c_s(x_2) \\
        \vdots   & \vdots   & \ddots & \vdots   \\
        c_1(x_p) & c_2(x_p) & \cdots & c_s(x_p)
    \end{bmatrix}.
\end{align*}

Introducing the RKHS methodology outlined in \Cref{subsec:rkhs}, the functions $c_{\gamma}$ in $\widetilde{\textbf{C}} = [c_1 \cdots c_s]$ for $\gamma \in [s]$ are chosen from a RKHS with kernel function $\mathcal{K}:I \times I \rightarrow \mathbb{R}$ where $I=\{x_1, \ldots,x_p\}\subset\mathbb{R}$; this yields a kernel matrix $\textbf{K}\in\mathbb{R}^{p \times p}$. Then, for the corresponding weight matrix $\textbf{W}\in\mathbb{R}^{p \times s}$, the functions $c_{\gamma}$ are described via
\begin{align*}
    c_{\gamma}(\cdot) = \sum_{k=1}^{p} w_{k\gamma} \mathcal{K}(\cdot,x_k),
\end{align*}
or $\widehat{\textbf{C}}_{I} = \textbf{K}\textbf{W}$ according to \Cref{eq:min_req_ls_sol}. As a result, for an evaluated input tensor $\widehat{\mathbfcal{T}}_{I}$, this yields the optimization problem
\begin{align}
    &\min_{\mathbfcal{G},\textbf{A},\textbf{B},\textbf{W}} &&\frac{1}{2} || \widehat{\mathbfcal{T}}_{I} - \mathbfcal{G} \bullet_1 \textbf{A} \bullet_2 \textbf{B} \bullet_3 \textbf{K}\textbf{W} ||^2 + \frac{\lambda}{2} ||\textbf{W}||_{\textbf{K}}^2,\label{eq:ftd_optimization_raw}
\end{align}
for $\mathbfcal{G}\in\mathbb{R}^{q \times r \times s}$, $\textbf{A}\in\mathbb{R}^{m \times q}$, $\textbf{B}\in\mathbb{R}^{n \times r}$, $\textbf{W}\in\mathbb{R}^{p \times s}$ and a regularization parameter $\lambda > 0$. Hereby, the regularization term stems from \Cref{eq:min_req_ls}. In the following, for fixed $I=\{x_1, \ldots,x_p\}$, we will denote $\widehat{\mathbfcal{T}}_{I}$ by $\widehat{\mathbfcal{T}}$.

\subsection{Optimization problem}\label{subsec:optimization_problem}
This section outlines how to solve the optimization problem (\labelcref{eq:ftd_optimization_raw}) independently for factor matrices of discrete modes $\textbf{A}$ and $\textbf{B}$, the weight matrix $\textbf{W}$ and the core tensor $\mathbfcal{G}$. In addition to the mathematical solution, we outline an algorithm to compute the solution in practice. 

\subsubsection{Solution for discrete modes}\label{par:solution_discrete_modes}
Without loss of generality, the goal is to find a solution to
\begin{align}
    \min_{\textbf{A}} \frac{1}{2} || \widehat{\mathbfcal{T}} - \mathbfcal{G} \bullet_1 \textbf{A} \bullet_2 \textbf{B} \bullet_3 \textbf{K}\textbf{W} ||^2 \label{eq:opt_discrete_modes} 
\end{align}
for all other factors fixed. By mode-$1$ unfolding, the solution of the problem is equivalent to
\begin{align*}
    \widehat{\mathbf{T}}_{(1)} = \textbf{A}(\mathbfcal{G} \bullet_2 \textbf{B} \bullet_3 \textbf{KW})_{(1)},
\end{align*}
according to \Cref{eq:tucker_unfolding1}. This formulation is computationally efficient because it avoids explicitly forming the Kronecker product. Analogously, with \Cref{eq:tucker_unfolding2}, for $\textbf{B}$ it follows
\begin{align*}
    \widehat{\mathbf{T}}_{(2)} = \textbf{B}(\mathbfcal{G} \bullet_1 \textbf{A} \bullet_3 \textbf{KW})_{(2)}.
\end{align*}
After solving the linear system, the solution is orthogonalized; for instance, by employing a SVD. In this case, the factor matrices $\textbf{A}$ and $\textbf{B}$ are replaced by their left leading singular vectors, while the remaining components of the decomposition are absorbed into the core tensor. Consequently, the factor matrices $\textbf{A}$ and $\textbf{B}$ can be treated as orthogonal in the subsequent steps.

\subsubsection{Solution for the continuous mode}\label{par:solution_continuous_mode}
The goal is to find a solution to
\begin{align*}
    \min_{\textbf{W}} \frac{1}{2} || \widehat{\mathbfcal{T}} - \mathbfcal{G} \bullet_1 \textbf{A} \bullet_2 \textbf{B} \bullet_3 \textbf{K}\textbf{W} ||^2 + \frac{\lambda}{2} ||\textbf{W}||_\textbf{K}^2 \quad \subjto \ \textbf{W}\in\mathbb{R}^{p \times s}
\end{align*}
for $\lambda>0$. Let $F:\mathbb{R}^{p \times s}\rightarrow\mathbb{R}$ denote the objective function dependent on $\mathbf{W}$
\begin{align*}
    F(\textbf{W}) = \frac{1}{2} || \widehat{\mathbfcal{T}} - \mathbfcal{G} \bullet_1 \textbf{A} \bullet_2 \textbf{B} \bullet_3 \textbf{K}\textbf{W} ||^2 + \frac{\lambda}{2} ||\textbf{W}||_\textbf{K}^2.
\end{align*}
Then, by unfolding the first summand in the third mode, setting $\textbf{N} = \textbf{G}_{(3)}\trans{(\textbf{B} \otimes \textbf{A})}$, and rewriting the objective with the trace operator yields
\begin{align*}
    &\frac{1}{2} || \widehat{\mathbf{T}}_{(3)} - \textbf{K}\textbf{W}\textbf{G}_{(3)}\trans{(\textbf{B} \otimes \textbf{A})} ||^2 + \frac{\lambda}{2} ||\textbf{W}||_\textbf{K}^2\\
    =& \frac{1}{2} \Tr\left( \trans{(\widehat{\mathbf{T}}_{(3)} - \textbf{K}\textbf{W}\textbf{N})} (\widehat{\mathbf{T}}_{(3)} - \textbf{K}\textbf{W}\textbf{N}) \right) + \frac{\lambda}{2} \Tr\left( \trans{\textbf{W}}\textbf{K}\textbf{W} \right)\\
    =& \frac{1}{2} \Tr\left( \trans{\widehat{\mathbf{T}}}_{(3)}\widehat{\mathbf{T}}_{(3)} - \trans{\widehat{\mathbf{T}}}_{(3)}\textbf{K}\textbf{W}\textbf{N} - \trans{(\textbf{K}\textbf{W}\textbf{N})} \widehat{\mathbf{T}}_{(3)} + \trans{(\textbf{K}\textbf{W}\textbf{N})} (\textbf{K}\textbf{W}\textbf{N}) + \lambda \trans{\textbf{W}}\textbf{K}\textbf{W} \right)\\
    =& \frac{1}{2} \Tr\left( \trans{\widehat{\mathbf{T}}}_{(3)}\widehat{\mathbf{T}}_{(3)} - 2\trans{\textbf{N}}\trans{\textbf{W}}\trans{\textbf{K}}\widehat{\mathbf{T}}_{(3)} + \trans{\textbf{N}}\trans{\textbf{W}}\trans{\textbf{K}}\textbf{K}\textbf{W}\textbf{N} + \lambda \trans{\textbf{W}}\textbf{K}\textbf{W} \right)\\
    =& \frac{1}{2} \Tr\left( \trans{\widehat{\mathbf{T}}}_{(3)}\widehat{\mathbf{T}}_{(3)} - 2\trans{\textbf{W}}\trans{\textbf{K}}\widehat{\mathbf{T}}_{(3)}\trans{\textbf{N}} + \trans{\textbf{W}}\trans{\textbf{K}}\textbf{K}\textbf{W}\textbf{N}\trans{\textbf{N}} + \lambda \trans{\textbf{W}}\textbf{K}\textbf{W} \right)
\end{align*}
Computing the gradient of $F$ with respect to $\textbf{W}$ yields
\begin{align*}
    \nabla_{\textbf{W}}F = -\textbf{K}\widehat{\mathbf{T}}_{(3)}\trans{\textbf{N}} + \textbf{K}\textbf{K}\textbf{W}\textbf{N}\trans{\textbf{N}} + \lambda\textbf{K}\textbf{W}
\end{align*}
for symmetric $\textbf{K}=\trans{\textbf{K}}$. By factoring out the positive definite $\textbf{K}$ and setting equal to $0$, this is equivalent to
\begin{align*}
    \textbf{K}\textbf{W}\textbf{N}\trans{\textbf{N}} + \lambda\textbf{W} = \widehat{\mathbf{T}}_{(3)}\trans{\textbf{N}}
\end{align*}
Due to the orthogonality of $\textbf{A}$ and $\textbf{B}$ it holds $\textbf{N}\trans{\textbf{N}} = \textbf{G}_{(3)}\trans{\textbf{G}}$. Hence, the resulting linear systems is
\begin{align*}
    \textbf{K}\textbf{W}\textbf{G}_{(3)}\trans{\textbf{G}}_{(3)} + \lambda\textbf{W} = \widehat{\mathbf{T}}_{(3)}\trans{\textbf{N}}
\end{align*}
solved for $\textbf{W}$. After vectorizing and using \Cref{eq:kron_ABCD} this is equivalent to
\begin{align*}
    ( \textbf{G}_{(3)}\trans{\textbf{G}}_{(3)} \otimes \textbf{K} + \lambda \textbf{I}_{sp} )\vecc(\textbf{W}) = \vecc(\widehat{\mathbf{T}}_{(3)}\trans{\textbf{N}})
\end{align*}
solved for $\vecc{(\textbf{W})}$. 

As for the discrete modes, the solution is orthogonalized; by employing a SVD $\textbf{C} = \textbf{K}\textbf{W} = \textbf{U}\bm{\Sigma}\trans{\textbf{V}}$. Then, $\textbf{C}$ is substituted by $\textbf{U}$ and $\textbf{W}$ is replaced by $\textbf{W}\textbf{V}\bm{\Sigma}^{-1}$. The remaining product $\bm{\Sigma}\trans{\bm{V}}$ is pushed to the core tensor, i.e., $\mathbfcal{G}\bullet_3\bm{\Sigma}\trans{\bm{V}}$.

\subsubsection{Solution for the core tensor}\label{par:solution_core_tensor}

With all factor matrices fixed, the core tensor is obtained from
\begin{align*}
    \min_{\mathbfcal{G}} \frac{1}{2}
    \big\| \widehat{\mathbfcal{T}} - \mathbfcal{G} \bullet_1 \textbf{A}
        \bullet_2 \textbf{B} \bullet_3 \textbf{K}\textbf{W} \big\|^2
    \qquad \subjto \ \mathbfcal{G}\in\mathbb{R}^{q \times r \times s},
\end{align*}
which vectorizes to the linear least squares problem
\begin{align*}
    \min_{\mathbfcal{G}} \frac{1}{2}
    \big\| \vecc(\widehat{\mathbfcal{T}})
        - (\textbf{K}\textbf{W} \otimes \textbf{B} \otimes \textbf{A})\,
          \vecc(\mathbfcal{G}) \big\|^2 .
\end{align*}
Due to the orthogonalization of $\textbf{A}$, $\textbf{B}$ and $\textbf{K}\textbf{W}$ in the previous steps and the mixed-product property of the Kronecker product \eqref{eq:kron_transpose}, $\textbf{K}\textbf{W} \otimes \textbf{B} \otimes \textbf{A}$ has orthonormal columns. Hence, the solution for $\mathbfcal{G}$ is $\vecc(\mathbfcal{G}) = \trans{(\textbf{K}\textbf{W} \otimes \textbf{B} \otimes \textbf{A})}\vecc(\widehat{\mathbfcal{T}})$, or
\begin{align*}
    \mathbfcal{G} = \widehat{\mathbfcal{T}} \bullet_1 \trans{\textbf{A}} \bullet_2 \trans{\textbf{B}} \bullet_3 \trans{(\textbf{K}\textbf{W})},
\end{align*}
in tensor notation.

\subsection{Algorithm}\label{subsec:algorithm}

\Cref{alg:ftd} shows the computation of the functional Tucker decomposition for a $3$-way tensor with two discrete and a continuous third mode. The orthogonalization step for the discrete modes, as explained in \Cref{par:solution_discrete_modes}, is based on the SVD. For instance, the factor matrix $\textbf{A}$ is decomposed according to $\textbf{A} = \textbf{U}_1\bm{\Sigma}_1\trans{\textbf{V}}_1$. Then, the factor matrix is updated to be $\textbf{U}_1$ while $\bm{\Sigma}_1\trans{\textbf{V}}_1$ is pushed to the core, i.e., $\mathbfcal{G} \gets \mathbfcal{G} \bullet_1 (\bm{\Sigma}_1 \trans{\textbf{V}}_1)$. The same holds for $\textbf{B}$. Furthermore, the placeholder function $\text{RESHAPE}$ reshapes vectors to matrices and matrices to tensors according to the inserted dimensions. Finally, a break condition is built into the algorithm based on the change of the relative approximation error $|\epsilon_t - \epsilon_{t-1}|$ between iterations $t-1$ and $t$. If the change is smaller than a pre-defined threshold $\tau$ multiplied by the norm of the input tensor, the algorithm terminates.

Each iteration in \Cref{alg:ftd} requires $\mathcal{O}(p^3s^3 + q^2m + r^2n + qrspn + qrspm + qrpmn)$ flops for an input tensor $\mathbfcal{T}\in\mathbb{R}^{m \times n \times p}$ and truncation parameters $q\leq m$, $r\leq n$ and $s\leq p$. Assuming $m,n < p$, this simplifies to $\mathcal{O}(p^3s^3)$ flops per iteration. Due to the $sp \times sp$ linear system solved in the continuous-mode block, this step is computationally the most expensive. The derivation of the computational cost can be found in \Cref{app:subsec:complexity}.

\begin{algorithm}
\caption{Functional Tucker decomposition for $3$-way tensor with continuous third mode}
\label{alg:ftd}
\begin{algorithmic}[1]
\REQUIRE{$\mathbfcal{T}\in\mathbb{R}^{m \times n \times p}$, $\textbf{K}\in\mathbb{R}^{p \times p}$, $\textsc{Maxiters}\in\mathbb{N}$, $\tau\in\mathbb{R}_{+}$}
\ENSURE{{$\{\mathbfcal{G},\textbf{A},\textbf{B},\textbf{W}\}$} such that $\mathbfcal{T} \approx \mathbfcal{G} \bullet_1 \textbf{A} \bullet_2 \textbf{B} \bullet_3 \textbf{K}\textbf{W}$}

\STATE{\text{Initialize } $\textbf{A}\in\mathbb{R}^{m \times q}$, $\textbf{B}\in\mathbb{R}^{n \times r}$, $\textbf{W}\in\mathbb{R}^{p \times s}$, $\mathbfcal{G}\in\mathbb{R}^{q \times r \times s}$}
\STATE{$\textbf{C} \gets \textbf{K}\textbf{W}$}
\FOR{$t=1,2,\ldots,\textsc{Maxiters}$}
    \STATE{}

    \STATE{$\textbf{A}_1 \gets$ \text{Solution of} $\textbf{A}_1 \left( \mathbfcal{G}\bullet_2\textbf{B}\bullet_3\textbf{K}\textbf{W} \right) = \textbf{T}_{(1)}$}
    \STATE{$\left[ \textbf{U}_1,\bm{\Sigma}_1,\textbf{V}_1 \right] \gets \text{SVD}(\textbf{A}_1)$}
    \STATE{$\textbf{A} \gets \textbf{U}_1$}
    \hfill Update 1st discrete factor matrix 
    \STATE{$\mathbfcal{G} \gets \mathbfcal{G} \bullet_1 (\bm{\Sigma}_1\trans{\textbf{V}_1})$}
    \hfill Update core tensor
    \STATE{}
    
    \STATE{$\textbf{B}_1 \gets$ \text{Solution of} $\textbf{B}_1 \left( \mathbfcal{G}\bullet_1\textbf{A}\bullet_3\textbf{K}\textbf{W} \right) = \textbf{T}_{(2)}$}
    \STATE{$\left[ \textbf{U}_2,\bm{\Sigma}_2,\textbf{V}_2 \right] \gets \text{SVD}(\textbf{B}_1)$}
    \STATE{$\textbf{B} \gets \textbf{U}_2$}
    \hfill Update 2nd discrete factor matrix
    \STATE{$\mathbfcal{G} \gets \mathbfcal{G} \bullet_2 (\bm{\Sigma}_2\trans{\textbf{V}_2})$}
    \hfill Update core tensor 
    \STATE{}
    
    \STATE{$\textbf{L} \gets \textbf{G}_{(3)}\trans{\textbf{G}_{(3)}} \otimes \mathbf{K} + \lambda \textbf{I}_{sp}$}
    \STATE{$\textbf{R} \gets \vecc(\textbf{T}_{(3)} (\textbf{B} \otimes \textbf{A})\trans{\textbf{G}_{(3)}})$}
    \STATE{ $\textbf{W}_{vec} \gets $ \text{Solution of} $\textbf{L}\textbf{W}_{vec} = \textbf{R}$}
    \STATE{$\textbf{W} \gets \text{RESHAPE}(\textbf{W}_{vec},[p \ s ])$}
    \STATE{$\left[ \textbf{U}_3,\bm{\Sigma}_3,\textbf{V}_3 \right] \gets \text{SVD}(\textbf{K}\textbf{W})$}
    \STATE{$\textbf{W} \gets \textbf{W}\textbf{V}_3\bm{\Sigma}_{3}^{-1}$}
    \STATE{$\textbf{C} \gets \textbf{U}_3$}
    \hfill Update continuous factor matrix 
    \STATE{$\mathbfcal{G} \gets \mathbfcal{G} \bullet_3 (\bm{\Sigma}_3\trans{\textbf{V}_3})$}
    \hfill Update core tensor     
    \STATE{}

    \STATE{$\mathbfcal{G} \gets \mathbfcal{T} \bullet_1 \trans{\textbf{A}} \bullet_2 \trans{\textbf{B}} \bullet_3 \trans{\textbf{C}}$}\hfill Update core tensor    
    \STATE{}
    
    \STATE $\epsilon_t \gets ||\mathbfcal{T} - \mathbfcal{G} \bullet_1 \textbf{A} \bullet_2 \textbf{B} \bullet_3 \textbf{C}|| \slash ||\mathbfcal{T}||$
    \hfill Check stop condition
    \IF{$t>1 \text{ and } |\epsilon_t - \epsilon_{t-1}| < \tau$}
        \STATE{\textbf{break}}
    \ENDIF
    \STATE{}
\ENDFOR
\RETURN{$\{\mathbfcal{G},\textbf{A},\textbf{B},\textbf{W}\}$}    
\end{algorithmic}
\end{algorithm}

\subsection{Reconstruction error under subspace transfer}\label{subsec:reconstr_err}

A theoretically sound argument for the subspace transfer presented in the experiments is provided by an upper bound of the reconstruction error under the transfer scenario. Once the FTD has been fitted on the design set $I=\{ x_1,\dots,x_p\}\subset\mathbb{R}$, its continuous factor can be evaluated on a second, in general distinct, transfer domain $\tilde I=\{\tilde x_1,\dots,\tilde x_{|\tilde I|}\}\subset\mathbb{R}$. Collecting the true and estimated functional columns sampled on $\tilde I$ into $\mathbf{C}^{*}_{\tilde I}=[\,c^{*}_1(\tilde I)\;\cdots\;c^{*}_s(\tilde I)]$ and $\widehat{\mathbf{C}}_{\tilde I}=[\,\hat c_1(\tilde I)\;\cdots\;\hat c_s(\tilde I)]$, respectively, the quantity of interest is the transfer-domain reconstruction error $\|\widehat{\mathbf{C}}_{\tilde I}-\mathbf{C}^{*}_{\tilde I}\|$. The following theorem bounds it in closed form. Moreover, it includes an expansion to the full tensor bound. The proof, together with supporting material on the workflow, is deferred to \Cref{app:subsec:reconstr_err}.

Before proceeding with the error bound, we state three assumptions for the fit obtained from \eqref{eq:ftd_optimization_raw}:
\begin{assumption}\label{ass:transfer}
\begin{enumerate}[label=(\Alph*), ref=\Alph*]
    \item\label{ass:kernel_matr_pd} The kernel matrix $\textbf{K}$ on $I$ is positive definite;
    \item\label{ass:core_unf_full_rank} the core unfolding $\textbf{G}_{(3)}\in\mathbb{R}^{s\times qr}$ has full row rank,
    so that $\textbf{M}:=\textbf{G}_{(3)}\trans{\textbf{G}}_{(3)}\in\mathbb{R}^{s\times s}$ is symmetric positive definite with eigenvalues $0<\mu_{1}\le\cdots\le\mu_{s}$;
    \item\label{ass:A_B_orth} the discrete factors $\textbf{A},\textbf{B}$ are orthonormal
    ($\trans{\textbf{A}}\textbf{A}=\textbf{I}_q$ and $\trans{\textbf{B}}\textbf{B}=\textbf{I}_r$,
    as enforced by the orthogonalization step of \Cref{par:solution_discrete_modes}), and
    the evaluated tensor is well specified and noiseless,
    \begin{align*}
        \widehat{\textbf{T}}_{(3)} = \textbf{C}_{I}^{*}\,\textbf{G}_{(3)}\,\trans{(\textbf{B}\otimes\textbf{A})}, \qquad
        \textbf{C}_{I}^{*}=[\,c_{1}^{*}(I)\;\cdots\;c_{s}^{*}(I)],
    \end{align*}
    i.e.\ $\widehat{\mathbfcal{T}}_I$ is generated exactly by the model with the true factors.
\end{enumerate}
\end{assumption}

In practice, \Cref{ass:kernel_matr_pd} holds for the Gaussian kernel with pairwise distinct points \cite{Wendland2004}. \Cref{ass:core_unf_full_rank} requires $s \le qr$, which we enforce through the choice of truncation parameters. The orthonormality in \Cref{ass:A_B_orth} is guaranteed by the orthogonalisation step of the algorithm, whereas the noiseless, well-specified generation of $\widehat{\mathbfcal{T}}_I$ is an idealisation adopted for the analysis. 

\begin{theorem}[Transfer reconstruction bound]\label{thm:transfer_bound}
Let \Cref{ass:kernel_matr_pd,ass:core_unf_full_rank,ass:A_B_orth} hold and let the continuous factor $\widehat{\mathbf{C}}$ be obtained from the inner update of \eqref{eq:ftd_optimization_raw} with FTD regularization parameter $\lambda_{\mathrm F}>0$. Then
\begin{equation}
    \big\|\widehat{\mathbf{C}}_{\tilde I}-\mathbf{C}^{*}_{\tilde I}\big\|
    \;\le\;
    \sqrt{|\tilde I|}\;\,
    P_{\mathcal{K},\,\lambda_{\mathrm F}/\mu_1}^{\max}(\tilde I,I)\;\,
    \|\mathbf{C}^{*}\|_{\mathcal{H}_\mathcal{K}},
    \label{eq:transfer_bound_main}
\end{equation}
where $\mu_1>0$ is the smallest eigenvalue of
$\mathbf{M}=\mathbf{G}_{(3)}\trans{\mathbf{G}}_{(3)}$, the aggregate RKHS norm is
$\|\mathbf{C}^{*}\|_{\mathcal{H}_\mathcal{K}}^2:=\sum_{\sigma=1}^{s}\|c^{*}_\sigma\|_{\mathcal{H}_{\mathcal{K}}}^2$, and,
with $\mathbf{k}(\tilde x)=\trans{(\mathcal{K}(\tilde x,x_1),\dots,\mathcal{K}(\tilde x,x_p))}$, the
regularized power function is
\begin{equation}
  P_{\mathcal{K},\lambda}(\tilde x,I)
  := \Big(\mathcal{K}(\tilde x,\tilde x)
     - \trans{\mathbf{k}(\tilde x)}(\mathbf{K}+\lambda\mathbf{I}_p)^{-1}\mathbf{k}(\tilde x)
     - \lambda\,\trans{\mathbf{k}(\tilde x)}(\mathbf{K}+\lambda\mathbf{I}_p)^{-2}\mathbf{k}(\tilde x)
     \Big)^{1/2},
  \label{eq:reg_power_main}
\end{equation}
with $P_{\mathcal{K},\lambda}^{\max}(\tilde I,I):=\max_{\tilde x\in\tilde I}P_{\mathcal{K},\lambda}(\tilde x,I)$.
Moreover, this bound lifts to the full tensor reconstruction error
\begin{equation}
    \big\|\widehat{\mathbfcal{T}}_{\tilde I}-\mathbfcal{T}^{*}_{\tilde I}\big\|
    \;\le\;
    \sqrt{|\tilde I|\,\mu_s}\;\,
    P_{\mathcal{K},\,\lambda_{\mathrm F}/\mu_1}^{\max}(\tilde I,I)\;\,
    \|\mathbf{C}^{*}\|_{\mathcal{H}_\mathcal{K}},
    \label{eq:transfer_bound_tensor}
\end{equation}
where $\widehat{\mathbfcal{T}}_{\tilde I}=[\![\mathbfcal{G};\mathbf{A},\mathbf{B},\widehat{\mathbf{C}}_{\tilde I}]\!]$, $\mathbfcal{T}^{*}_{\tilde I}=[\![\mathbfcal{G};\mathbf{A},\mathbf{B},\mathbf{C}^{*}_{\tilde I}]\!]$, and $\mu_s$ is the largest eigenvalue of $\mathbf{M}$.
\end{theorem}
The bound \eqref{eq:transfer_bound_main} factorizes the matrix-level transfer error into three interpretable quantities. First, the cardinality factor $\sqrt{|\tilde I|}$ accounting for the number of transfer design points. Second, the purely geometric regularized power function $P_{\mathcal{K},\lambda}^{\max}$ which depends only on the kernel and on the configuration of $\tilde I$ relative to $I$; it measures how well the training design points $I$ \textit{cover} the transfer points $\tilde I$ relative to the kernel's native geometry, decreasing to the interpolation power function as $\lambda\downarrow0$ and vanishing when $\tilde I\subseteq I$ in that limit. The effective regularization $\lambda_{\mathrm F}/\mu_1$ couples the error to the conditioning of the core: $P_{\mathcal{K},\lambda}$ is monotonically increasing in $\lambda$, so an ill-conditioned core (small $\mu_1$) inflates the worst-case regularization and loosens the guarantee, whereas a well-conditioned core tightens it. And third, the aggregate RKHS norm $\|\mathbf{C}^{*}\|_{\mathcal{H}_\mathcal{K}}$ of the true continuous factor which measures the RKHS complexity (smoothness) of the true continuous factors and is the only target-dependent term. 
The tensor-level bound \eqref{eq:transfer_bound_tensor} preserves this structure: the power function and the RKHS-complexity term carry over unchanged, as the multilinear Tucker structure propagates the continuous-mode error through the orthonormal discrete factors without amplification. The lift adds only the spectral prefactor $\sqrt{\mu_s}$, the worst-case gain with which the core amplifies a continuous-factor
perturbation as it is mapped into the tensor. In the noiseless setting, the remaining error arises from the rank truncations of the discrete modes; further details on unconstrained Tucker approximation can be found in \cite{DeLathauwerDeMoorVandewalle2000_2}.

\section{Experimental results}\label{sec:experiments}

This section demonstrates the capabilities of adaptive subspace modeling with the functional Tucker decomposition in three experiments. First, a semi-synthetic dataset, built on the traditional task of digit classification, is considered, while the second and third experiment turn to real-world data in the realm of classification of multivariate time-series and hyperspectral images. By stacking samples in an additional mode, all three examples involve $4$-way tensors. Since the datasets contain only one continuous mode (with the remaining modes being discrete), the theoretical framework presented above remains applicable.

\subsection{Semi-synthetic data}\label{subsec:exp_synthetic}

\subsubsection{Dataset} 
The first experiment examines a semi-synthetic dataset which builds upon the US postal service (USPS) database \cite{LeCunEtAl1989}. The USPS dataset consists of $16 \times 16$ images of digits from $0$ to $9$ stemming from ZIP code scans of US Postal mail envelopes and -- among others -- was used to showcase the classification with the HOSVD in \cite{SavasElden2007}. 

\begin{figure}[tbhp]
\centering
\includegraphics[width=0.95\linewidth]{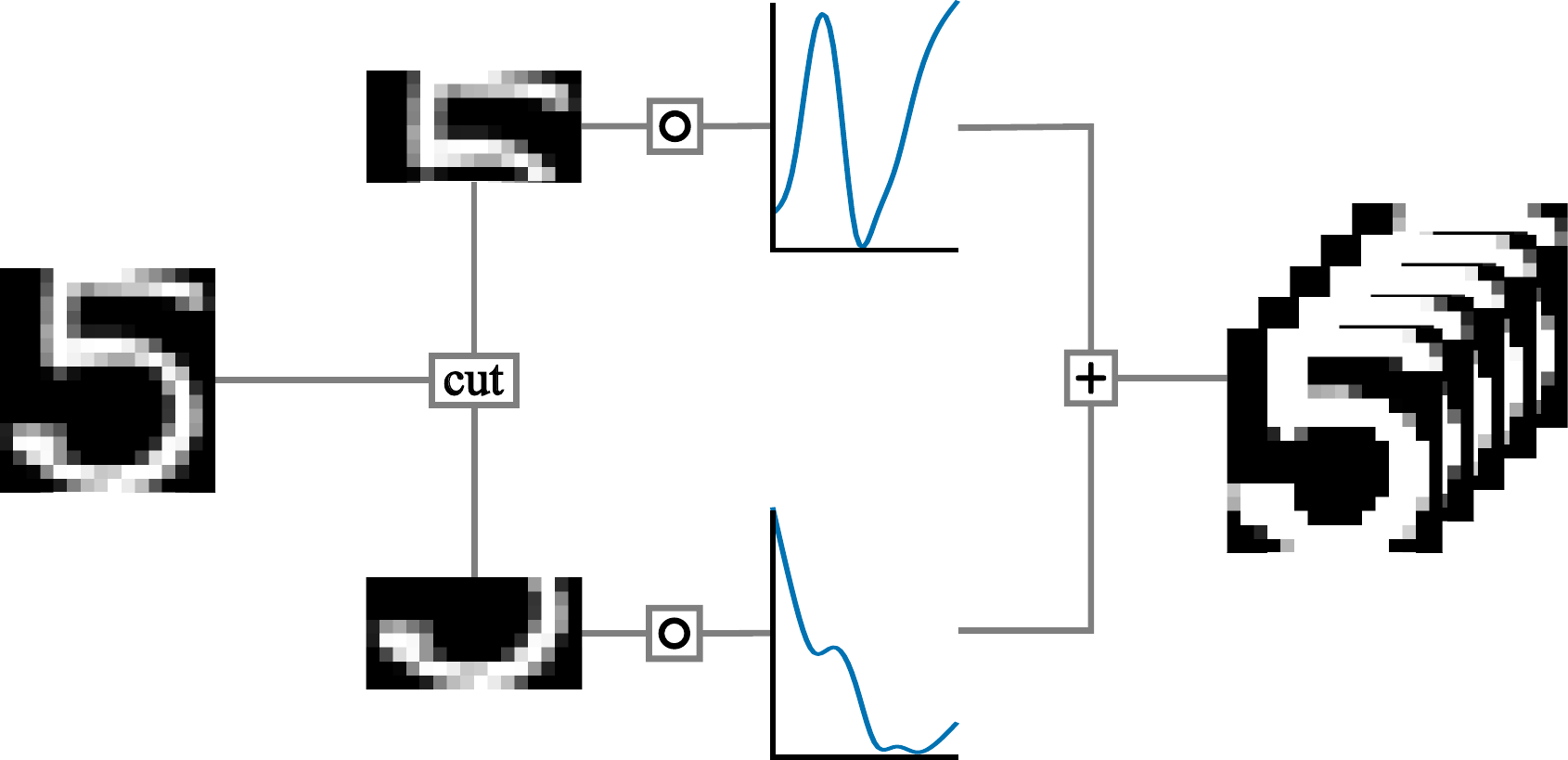}
\caption{Expanding a digit sample by a continuous mode based on smoothing splines.}
\label{fig:ex1_construction}
\end{figure}

By introducing an additional mode based on continuous functions, we can leverage adaptive subspace modeling within the FTD and analyze the adapted digit classification problem in this extended setting. To this end, each digit is split into two parts. These sub‑parts are then individually multiplied via an outer product with class‑specific function evaluations and recombined afterwards. The process of expanding a digit by a continuous mode is illustrated in \Cref{fig:ex1_construction}.

From a technical perspective, the functions associated with class $c$ are represented by two evaluation vectors $s_{c,\text{upper}}\in\mathbb{R}^{p}$ and $s_{c,\text{lower}}\in\mathbb{R}^{p}$. For a sample $\textbf{Y}_c\in\mathbb{R}^{16 \times 16}$ of class $c$, the tensorized sample $\mathbfcal{Y}_c\in\mathbb{R}^{16 \times 16 \times p}$ is constructed as 
\begin{align*}
    \mathbfcal{Y}_c = \textbf{Y}_c^{\text{lower}} \circ s_{c,\text{lower}} + \textbf{Y}_c^{\text{upper}} \circ s_{c,\text{upper}}.
\end{align*} 
Here, $\textbf{Y}_c^{\text{lower}}$ denotes the sample with its upper half set to zero, while $\textbf{Y}_c^{\text{upper}}$ corresponds to the image with its lower half set to zero. By choosing the vectors $s_{c,\text{lower}}$ and $s_{c,\text{upper}}$ to be linearly independent, the mode-$3$ rank of $\mathbfcal{Y}_c$ is guaranteed to be $2$. Repeating this process for all samples of a certain digit class $c\in\{0,1,\ldots,9\}$, the final class objects are obtained by stacking samples of the same class in the first mode, yielding a tensor $\mathbfcal{X}_c \in \mathbb{R}^{N_c \times 16 \times 16 \times p}$. Here, $N_c$ denotes the number of samples for the respective class. We construct $s_{c,\text{lower}}$ and $s_{c,\text{upper}}$ by fitting smoothing splines to $10$ uniform random samples from $(1,10)$ and evaluating them at $p=50$ points.

In the following, we will use the semi-synthetic digit dataset to illustrate the classification in the context of a changed domain after training. To this end, two different sampling domains are considered in training and test. Finally, the goal is to bridge both stages by application of adaptive subspace modeling with the FTD.

With the purpose of representing two different domains for training and test, the available set of $p$ sampling points is split. Specifically, the training is based on every fourth sampling point in the continuous mode, i.e., the training employs the tensor built of slices $\mathbfcal{X}_c(:,:,:,i)$ for $i\in\{1,5,9,\ldots,45,49\}$. Then, the computation of the FTD, based on this reduced domain, enables the interpolation of missing sampling points by evaluating the complete set of $p=50$ sampling points. 

\begin{figure}[tbhp]
\centering
\includegraphics[width=\linewidth]{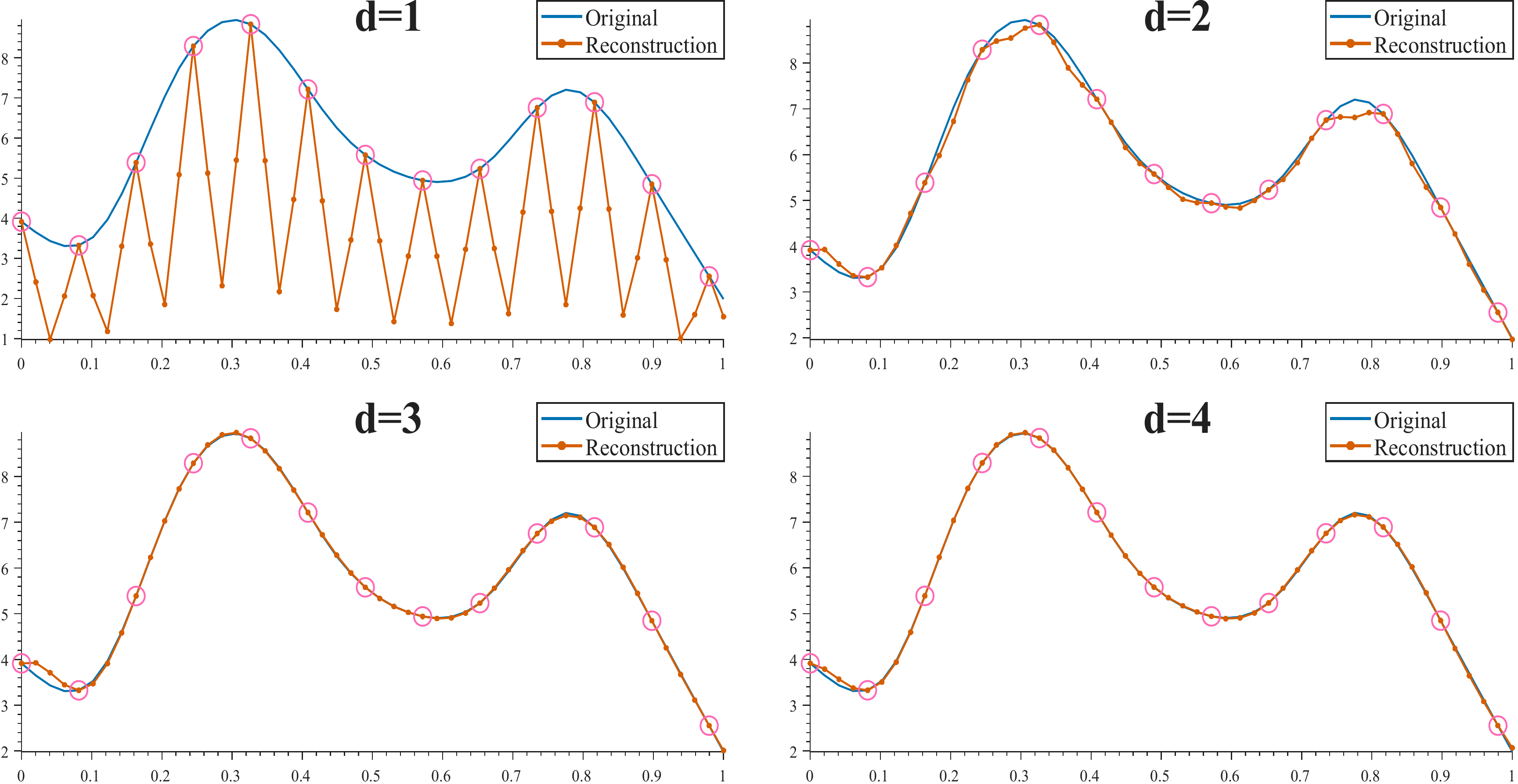}
\caption{Original and reconstruction of $\mathbfcal{X}_5(1,5,5,:)$.}
\label{fig:ex1_reconstr_first0_5_5}
\end{figure}

For illustration purposes, the reconstruction of $\mathbfcal{X}_5(1,5,5,:)$, i.e., the continuous mode of $\mathbfcal{X}$ for the first sample of class $5$ and the pixel at position $(5,5)$, is displayed in \Cref{fig:ex1_reconstr_first0_5_5}. In particular, the reconstruction is compared for Gaussian kernel parameters $d\in\{1,2,3,4\}$. The selection of the Gaussian kernel parameter is relevant for achieving an accurate reconstruction. Furthermore, although the reconstructed curves provide a satisfactory approximation of the true function for suitable $d$, the distribution of sampling points inherently limits the best possible approximation. 

\subsubsection{Classification} 
After having illustrated the interpolation capacity of the FTD, the functional approximation is used for inter-domain classification. Here, we will use $85\%$ of the data for training while the remaining $15\%$ are reserved for testing. As before, the training data is constructed by restricting the sampling points to every fourth index, before the FTD yields a corresponding factorization. In a next step, the goal is to run the classification on data with a different sampling domain. Therefore, the test data is adjusted to only contain every available sample point between indices $1$ and $13$, requiring a transfer between discretizations. By employing the same number of sample points for training and test, we make sure the FTD can be compared to the HOSVD and vectorized baseline methods. In order to incorporate the discretization shift, after obtaining the functional Tucker decomposition, the reconstruction of the class tensor is computed by employing the kernel matrix with corresponding domain, i.e., the set of sample points between indices $1$ and $13$. This yields a reconstruction of the class tensor in the new domain. Then, the HOSVD of the reconstruction enables the HOSVD-driven classification approach described in \Cref{subsec:class_hosvd}. This approach is repeated for all classes and, finally, an unknown sample can be classified by comparing it to all available bases and choosing the best according to the lowest residual.

The truncation parameters are set to $5$ in the spatial mode and $2$ in the continuous mode, according to existing knowledge from \cite{SavasElden2007} and construction of the continuous mode. Moreover, the FTD is run with Gaussian kernel parameter $d=4$ following the comparison above. 
 
\begin{figure}[tbhp]
\centering
\includegraphics[width=\linewidth]{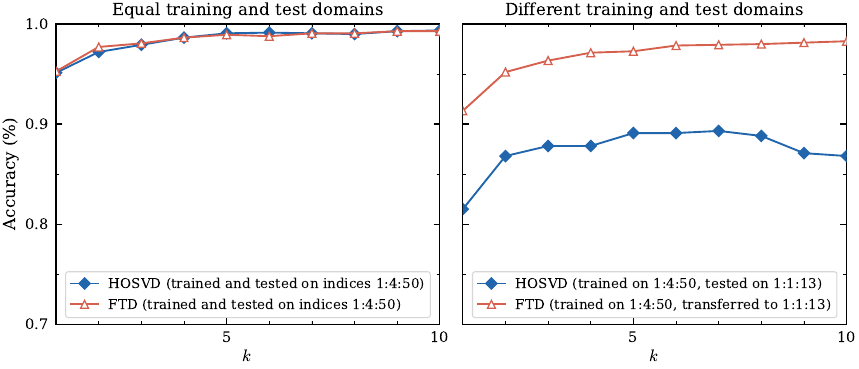}
\caption{Success rates of HOSVD- and FTD-driven classification for equal (l.h.s.) and different (r.h.s.) training and test domains for the semi-synthetic digit data. While the FTD and the HOSVD achieve comparable performance when the sampling points during training and test match, a noticeable performance gap emerges when different sampling points are used.}
\label{fig:ex1_successrates_lambda_1_d_4}
\end{figure}

Finally, \Cref{fig:ex1_successrates_lambda_1_d_4} illustrates the classification results with and without the domain change for the HOSVD model and the FTD model. In particular, \Cref{fig:ex1_successrates_lambda_1_d_4} indicates the success rates for different choices of truncation in the sample mode. Individually, the figures compare the corresponding metrics with equal sets of sampling points for training and test on the left hand side, while the right hand side is based on different training and test domains. While equal sampling points used in training and test yield similar success rates for both the HOSVD and FTD, different sampling points cause the metric to decline for the HOSVD which does not include the subspace transfer. \Cref{app:fig:ex1_macrof1scores_lambda_1_d_4} in the appendix reports an analogous behavior for macro F1 scores.

\Cref{tab:results_semisynthetic} lists the results in comparison to other vector- and tensor-based models. In the same-domain setting, all methods approach saturation: the MLP attains the highest accuracy ($99.96\%$) and macro F1 ($99.96\%$), closely followed by the SVM ($99.93\%$) and $k$NN ($99.50\%$), while HOSVD and FTD remain competitive with $99.35\%$ ($k=10$) and $99.28$ ($k=9$), confirming that the task is well-solved in the source domain across all model families. The transferred-domain results reveal a divergence in generalization behavior. The MLP suffers the most severe degradation, collapsing to $62.47\%$ accuracy, and the SVM drops to $70.75\%$. $k$NN is comparatively more robust, retaining $91.33\%$ accuracy. HOSVD also degrades noticeably, declining to $89.32\%$ ($k{=}7$) and $86.81\%$ ($k{=}10$) despite its competitive same-domain performance, confirming that a fixed source-domain subspace does not transfer reliably under domain shift. In contrast, FTD achieves $98.28\%$ accuracy and $98.16\%$ macro F1 ($k{=}10$), surpassing all baselines by a substantial margin. Moreover, the FTD performs on par with the oracle configurations which are HOSVD models trained on the test domain. Here, the marginal better performance of the FTD might be an artifact of the very limited transferred domain and is unlikely to be practically significant. Overall, these results suggest that the functional representation learned from the source domain captures structure that transfers reliably across domains, leveling with the performance ceiling achievable by a fully supervised oracle without access to any target-domain labels.
\begin{table}[tbhp]
    \centering
    \caption{\parindent=0pt Classification results on the semi-synthetic digit dataset. Training domain: indices \texttt{1:4:50}. Same-domain test: indices \texttt{1:4:50}. Transferred domain test: indices \texttt{1:1:13}. Best results are bold.}
    \label{tab:results_semisynthetic}
    \begin{threeparttable}                 
    \small
    \setlength{\tabcolsep}{5pt}    
    \begin{tabular}{@{}lcccc@{}}
        \toprule
        & \multicolumn{2}{c}{Same domain}
        & \multicolumn{2}{c}{Transferred domain} \\
        \cmidrule(lr){2-3}\cmidrule(lr){4-5}
        Model
          & Accuracy (\%)
          & Macro F1 (\%)
          & Accuracy (\%)
          & Macro F1 (\%) \\
        \midrule
        \rowcolor{rowgray}
        \multicolumn{5}{@{}l}{%
          \textit{Supervised baselines (vectorised input)}} \\
        $k$NN ($k{=}1$)\tnote{$\ddagger$}
          & $99.50$ & $99.43$
          & $91.33$ & $91.12$ \\
        MLP\tnote{$*$}
          & \bm{$99.96 \pm 0.00$} & \bm{$99.96 \pm 0.00$}
          & $62.68 \pm 0.04$ & $55.70 \pm 0.05$ \\
        SVM\tnote{$\dagger$}
          & $99.93$ & $99.93$
          & $70.75$ & $62.48$ \\
        \rowcolor{rowgray}
        \multicolumn{5}{@{}l}{%
          \textit{Tensor subspace baseline}} \\
        HOSVD ($k{=}7$)\tnote{$^\circ$}
          & $99.07$ & $98.98$
          & $89.32$ & $86.95$ \\
        HOSVD ($k{=}10$)\tnote{$^\circ$}
          & $99.35$ & $99.35$
          & $86.81$ & $82.84$ \\
        \rowcolor{rowgray}
        \multicolumn{5}{@{}l}{%
          \textit{Our method}} \\
        FTD ($k{=}9$)\tnote{$\S$}
          & $99.28$ & $99.25$
          & $98.14$ & $98.01$\\
        FTD ($k{=}10$)\tnote{$\S$}
          & $99.28$ & $99.25$
          & \bm{$98.28$} & \bm{$98.16$}\\
        \rowcolor{rowgray}
        \multicolumn{5}{@{}l}{%
          \textit{Oracle models}} \\
        HOSVD-Oracle ($k{=}7$)\tnote{$\P$}
          & \multicolumn{1}{c}{---} & \multicolumn{1}{c}{---}
          & $97.99$ & $97.83$ \\
        HOSVD-Oracle ($k{=}9$)\tnote{$\P$}
          & \multicolumn{1}{c}{---} & \multicolumn{1}{c}{---}
          & $97.78$ & $97.62$ \\
        HOSVD-Oracle ($k{=}10$)\tnote{$\P$}
          & \multicolumn{1}{c}{---} & \multicolumn{1}{c}{---}
          & $97.92$ & $97.82$ \\
        \bottomrule
    \end{tabular}
    \begin{tablenotes}[flushleft]\footnotesize
    \item[$*$] MLP: hidden layers $(50, 50)$, tanh activation, $\alpha = 0.0001$, determined by grid search. Applied to vectorized slices from the training domain; no subspace transfer is performed at test time. The results are mean $\pm$ std over 10 independent runs.
    \item[$\dagger$] SVM: RBF kernel, $\gamma = 1/(n_{\mathrm{features}} \cdot \mathrm{var}(X))$. Applied to vectorized slices from the training domain; no subspace transfer is performed at test time.
    \item[$\ddagger$] $k$NN: Euclidean distance on vectorised slices, $k$ selected by 5-fold cross-validation on training data. No subspace transfer is performed at test time.
    \item[$^\circ$] HOSVD: Tucker ranks set equal to $(5,5,2)$ (see \Cref{subsec:exp_synthetic}).
    \item[$\S$] FTD: regularization parameter $\lambda=1$ and Gaussian kernel bandwidth $d=4$ selected by cross-validation; Tucker ranks set equal to $(5,5,2)$ (see \Cref{subsec:exp_synthetic}).
    \item[$\P$] Oracle: model trained and evaluated entirely within the transfer test domain. Represents the performance under no domain shift. Tucker ranks set equal to $(5,5,2)$.
    \end{tablenotes}
    \end{threeparttable}
\end{table}

\subsection{Multivariate time-series data: Trip records of NYC taxis in 2024}\label{subsec:exp_ts}

\subsubsection{Dataset}
\begin{figure}[tbhp]
\centering
\includegraphics[width=\linewidth]{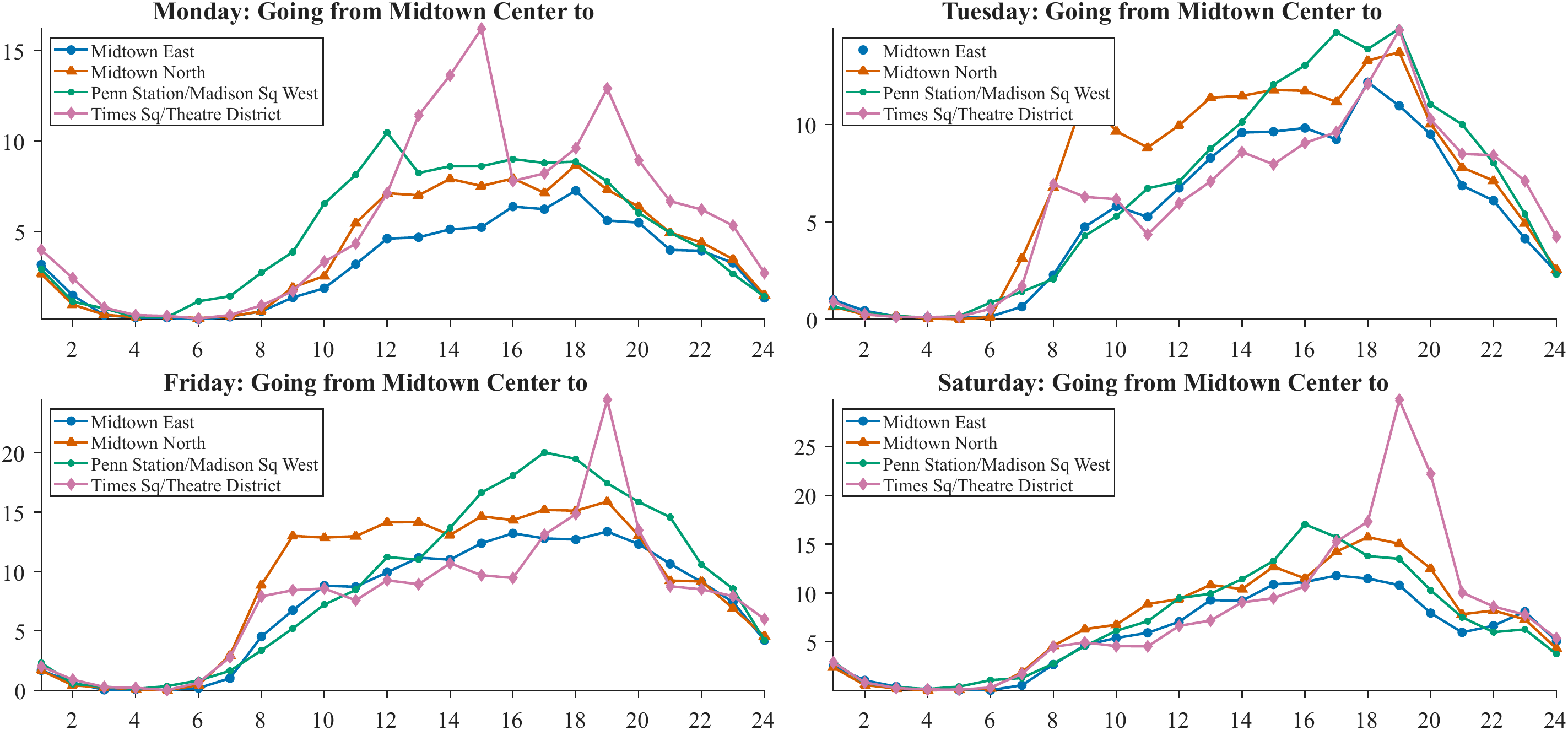}
\caption{Yearly average of NYC taxi trip records per time of the day going from Midtown Center (Location ID $161$) to Midtown East (Location ID $162$), Midtown North (Location ID $163$), Penn Station/Madison Sq West (Location ID $186$) and Times Sq/Theatre District (Location ID $230$) on Mondays, Tuesdays, Fridays and Saturdays.}
\label{fig:ex2_illustration}
\end{figure}

Since 2009, the NYC Taxi and Limousine Commission (TLC) provides monthly taxi trip records capturing various fields such as pickup and drop-off dates, times and locations \cite{NYCTaxiDataset2024}. The following analysis considers NYC taxi trip records from 2024, restraining the dataset to relevant trips that either begin or end at one of the locations listed in \Cref{tab:NYCTaxi_pu_and_do_locations}. 
\begin{table}[htbp]
\footnotesize
\caption{Relevant subset of NYC taxi pickup and drop-off locations.}\label{tab:NYCTaxi_pu_and_do_locations}
\begin{center}
    \begin{tabular}{|c|l|c|}\hline
        \textbf{Index} & \textbf{Zone} & \textbf{Location ID} \\
        \hline
        1 & Midtown Center & 161\\
        2 & Midtown East & 162\\
        3 & Midtown North & 163\\
        4 & Penn Station/Madison Sq West & 186\\
        5 & Times Sq/Theatre District & 230\\
        \hline
    \end{tabular}
\end{center}
\end{table}
The data is aggregated to hourly counts of trip records between two locations, split by day of the week. Hence, for each class $c\in\{ \text{Monday}, \text{Tuesday}, \ldots, \text{Sunday} \}$, a $4$-way tensor $\mathbfcal{X}_c \in \mathbb{R}^{N_c \times 5 \times 5 \times 24}$ covering $5$ pickup and drop-off locations over $24$ hours is considered. Hereby, the number of samples in class $c$ is denoted by $N_c$. For instance, the measurement $\mathbfcal{X}_{\text{Tuesday}}(4,2,3,9)=5$ indicates that $5$ trips were recorded from Midtown East to Midtown North on the fourth Tuesday of 2024 between 8am and 9am. For illustration purposes, \Cref{fig:ex2_illustration} presents yearly averages per time of the day for Monday, Tuesday, Friday and Saturday going from Midtown Center (Location ID $161$) to the other locations.

In a consistent approach to the previous experiment, the goal is to run the classification with a change of sampling points. Hence, the training data is based on every second sampling point in the continuous mode, i.e., the training uses the tensor built of slices $\mathbfcal{X}_c(:,:,:,i)$ for $i\in\{1,3,5,\ldots,21,23\}$. The goal is to run the classification on a different set of design points to illustrate the adaptive subspace modeling. In this case, the domain for testing is set to be every hour between $8$ and $19$.

\subsubsection{Classification} 

Before the subspace transfer, in a first step, the overall number of $N=348$ samples is partitioned into subsets for training + validation ($85\%$) and test ($15\%$). Without existing knowledge of the dataset, $5$-fold cross validation is executed on the training + validation data to determine optimal ranks for the classification. The validation step runs the classification based on the HOSVD instead of FTD to ensure that the factorization remains independent of the specific discretization shift under consideration. 

Resulting heatmaps for different truncation strategies in the sample mode, the modes for pickup and drop-off locations, and the time mode, are displayed in the appendix in \Cref{app:fig:ex2_ny_taxi_data_hosvd_cross_validation_k_start_loc} and \Cref{app:fig:ex2_ny_taxi_data_hosvd_cross_validation_k_end_loc}. Investigating the heatmaps, the validation suggests ranks of $5$ and $3$ for the pickup and drop-off location, respectively. A rank of $5$ is proposed for the time mode. Moreover, based on experimental results, the Gaussian kernel was used with parameter $d=1$ and the regularization parameter was set to $\lambda=1$.

The resulting classification accuracies are displayed in \Cref{fig:ex2_successrates_lambda_1_d_1}. Again, \Cref{app:fig:ex2_macrof1scores_lambda_1_d_1} in the appendix reports the curves for macro F1 scores. The results both in terms of the classification accuracy, as well as the macro F1 score indicate a successful transfer of the domain when comparing the metrics of the HOSVD and the FTD on the right-hand side of the figures. The variation of the results between the HOSVD and FTD in the context of equal training and test domains might be explained by a combination of the small sample size and the regularized optimization procedure of the FTD.
\begin{figure}[tbhp]
\centering
\includegraphics[width=\linewidth]{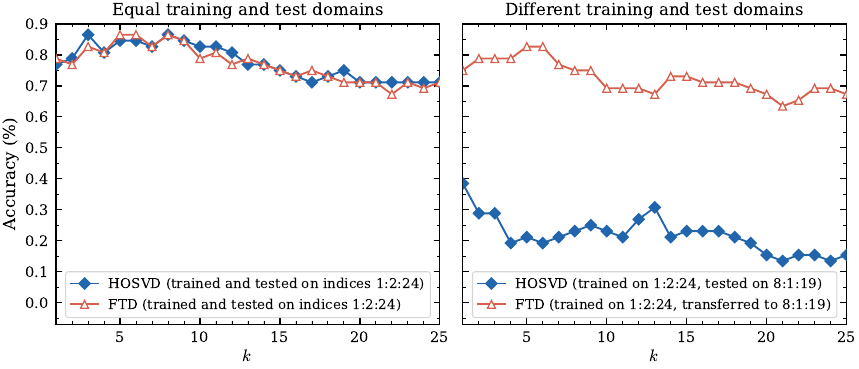}
\caption{Success rates of HOSVD- and FTD-driven classification for equal (l.h.s.) and different (r.h.s.) training and test domains for the NYC taxi trip data. On the left-hand side, classification accuracies of the FTD and the HOSVD-based model are similar, both varying between $60\%$ and $80\%$ for increasing $k$. On the right-hand side, however, a clear gap emerges due to the subspace transfer: while the FTD still achieves an accuracy of up to $82.69\%$ ($k{=}5)$, the HOSVD-based classification drops to a best-performing accuracy of $38.46\%$ ($k{=}1$).}
\label{fig:ex2_successrates_lambda_1_d_1}
\end{figure}

\Cref{tab:results_ny_taxi_data} lists the results in comparison to other vector- and tensor-based models. In the same-domain setting, the tensor subspace methods are competitive with or superior to the vectorized baselines: HOSVD and FTD with $k{=}6$ attain the highest same-domain accuracy of $82.69\%$. In the transferred-domain results all vectorized baselines suffer severe degradation: the MLP collapses to $23.85\%$ accuracy, the SVM drops to $40.38\%$, and $k$NN falls to $51.92\%$. Similarly, HOSVD degrades substantially from its strong same-domain performance to a maximum of $38.46\%$ under domain shift ($k{=}1$), indicating that a fixed subspace trained on the source domain does not transfer reliably. In contrast, FTD achieves $82.69\%$ accuracy and $82.21\%$ macro F1 on the transferred domain ($k{=}5$). Notably, the FTD attains accuracy on par with the HOSVD-Oracle across all truncation choices $k\in\{1,3,4,6\}$, despite the oracle's direct access to the transferred domain. In this case, the subspace transfer is essentially as good as retraining on the target domain.

\begin{table}[tbhp]
    \centering
    \caption{\parindent=0pt Classification results on the NCY taxi trip dataset. Training domain: indices \texttt{1:2:24}. Same-domain test: indices \texttt{1:2:24}. Transferred domain test: indices \texttt{8:1:19}. Best results are bold.}
    \label{tab:results_ny_taxi_data}
    \begin{threeparttable}                 
    \small
    \setlength{\tabcolsep}{5pt}    
    \begin{tabular}{@{}lcccc@{}}
        \toprule
        & \multicolumn{2}{c}{Same domain}
        & \multicolumn{2}{c}{Transferred domain} \\
        \cmidrule(lr){2-3}\cmidrule(lr){4-5}
        Model
          & Accuracy (\%)
          & Macro F1 (\%)
          & Accuracy (\%)
          & Macro F1 (\%) \\
        \midrule
        \rowcolor{rowgray}
        \multicolumn{5}{@{}l}{%
          \textit{Supervised baselines (vectorised input)}} \\
        $k$NN ($k{=}15$)\tnote{$\ddagger$}
          & $73.08$ & $71.50$
          & $51.92$ & $48.88$ \\
        MLP\tnote{$*$}
          & $74.07 \pm 0.06$ & $74.36 \pm 0.05$
          & $23.85 \pm 0.03$ & $17.38 \pm 0.03$ \\
        SVM\tnote{$\dagger$}
          & $80.77$ & $81.21$
          & $40.38$ & $34.32$ \\
        \rowcolor{rowgray}
        \multicolumn{5}{@{}l}{%
          \textit{Tensor subspace baseline}} \\
        HOSVD ($k{=}1$)\tnote{$^\circ$}
          & $76.92$ & $77.65$
          & $38.46$ & $34.56$ \\
        HOSVD ($k{=}3$)\tnote{$^\circ$}
          & $\bm{86.54}$ & $\bm{86.90}$
          & $28.85$ & $27.25$ \\
        \rowcolor{rowgray}
        \multicolumn{5}{@{}l}{%
          \textit{Our method}} \\
        FTD ($k{=}5$)\tnote{$\S$}
          & $\bm{86.54}$ & $86.89$
          & $\bm{82.69}$ & $82.21$\\
        \rowcolor{rowgray}
        \multicolumn{5}{@{}l}{%
          \textit{Oracle models}} \\
        HOSVD-Oracle ($k{=}1$)\tnote{$\P$}
          & \multicolumn{1}{c}{---} & \multicolumn{1}{c}{---}
          & $78.85$ & $78.84$ \\
        HOSVD-Oracle ($k{=}3$)\tnote{$\P$}
          & \multicolumn{1}{c}{---} & \multicolumn{1}{c}{---}
          & $80.77$ & $80.81$ \\    
        HOSVD-Oracle ($k{=}4$)\tnote{$\P$}
          & \multicolumn{1}{c}{---} & \multicolumn{1}{c}{---}
          & $\bm{82.69}$ & $\bm{82.70}$ \\
        HOSVD-Oracle ($k{=}5$)\tnote{$\P$}
          & \multicolumn{1}{c}{---} & \multicolumn{1}{c}{---}
          & $80.77$ & $80.83$ \\
        \bottomrule
    \end{tabular}
    \begin{tablenotes}[flushleft]\footnotesize
    \item[$*$] MLP: hidden layers $(50,)$, tanh activation, $\alpha = 0.001$, determined by grid search. Applied to vectorized slices from the training domain; no subspace transfer is performed at test time. The results are mean $\pm$ std over 10 independent runs.
    \item[$\dagger$] SVM: RBF kernel, $\gamma = 1/(n_{\mathrm{features}} \cdot \mathrm{var}(X))$. Applied to vectorized slices from the training domain; no subspace transfer is performed at test time.
    \item[$\ddagger$] $k$NN: Euclidean distance on vectorised slices, $k$ selected by 5-fold cross-validation on training data. No subspace transfer is performed at test time.
    \item[$^\circ$] HOSVD: Tucker ranks selected via HOSVD-based 5-fold cross-validation set to $(5,3,5)$ (see \Cref{subsec:exp_ts}).
    \item[$\S$] FTD: Tucker ranks selected via HOSVD-based 5-fold cross-validation set to $(5,3,5)$; regularization parameter and Gaussian kernel bandwidth set to $\lambda=1$ and $d=1$ (see \Cref{subsec:exp_ts}).
    \item[$\P$] Oracle: model trained and evaluated entirely within the transfer test domain. Represents the performance under no domain shift. Tucker ranks set equal to $(5,3,5)$.
    \end{tablenotes}
    \end{threeparttable}
\end{table}

\subsection{Hyperspectral image data: Hyperspectral images of mangos}\label{subsec:exp_hsi}

\subsubsection{Dataset}

In this final study, we address the classification of close-range hyperspectral image (HSI) data under a discretization shift scenario. Here, the change of domain is represented by different wavelengths present in the data. This situation mirrors the real-world use-case of two different camera setups. Overall, the dataset consists of hyperspectral images of mangos categorized into three ripeness classes: \textbf{UNRIPE}, \textbf{RIPE} and \textbf{OVERRIPE} \cite{VargaMakowskiZell2021}. Images were recorded using a Specim FX 10 camera, resulting in $224$ channels in a spectral range from $400$ to $1000$ nm and a spatial dimension of $64 \times 64$. Moreover, the dataset consists of $121$ samples in total which splits up into $39$, $52$ and $30$ instances of classes \textbf{UNRIPE}, \textbf{RIPE} and \textbf{OVERRIPE}, respectively.
As before, the goal is to run the classification with a change of domain. Therefore, let the training domain be defined by every fourth available sampling value in the spectral mode, while the classification results are inferred on data consisting of every sampling value between indices $100$ and $155$ which translates to the spectral range from $604.9250$ to $714.3090$ nm.

\subsubsection{Classification} 
Truncation parameters were determined by stratified validation comparing the classification accuracies yielded by the HOSVD-based classification approach. In particular, the truncation in the sample mode, the spatial mode and the spectral mode was tested on $85\%$ of all available samples. Hereby, due to a symmetry argument, the spatial rank was set equal for both spatial dimensions. Resulting heatmaps for different truncation strategies in the sample mode, the spatial modes and the spectral mode are displayed in the appendix in \Cref{app:fig:ex3_hsi_mango_data_hosvd_cross_validation,}. Evaluating the heatmaps, the validation suggests a a spatial rank of $5$ and a spectral rank of $14$. Moreover, based on experimental results, the Gaussian kernel was used with parameter $d=1$ and the regularization parameter was set to $\lambda=1$.

\begin{figure}[tbhp]
\centering
\includegraphics[width=\linewidth]{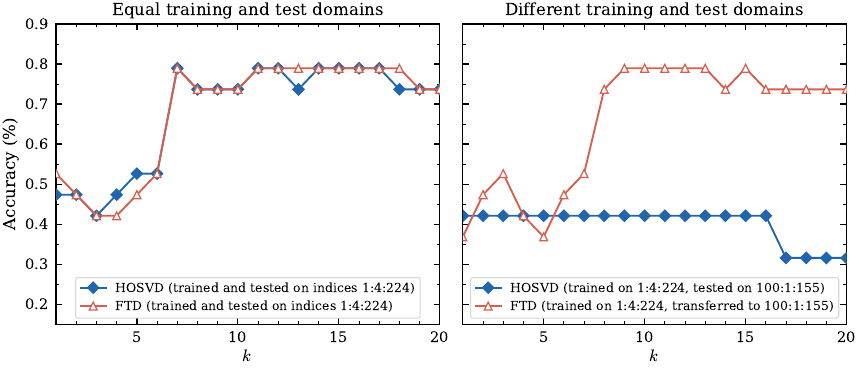}
\caption{Success rates of HOSVD- and FTD-driven classification for equal (l.h.s.) and different (r.h.s.) training and test domains for the mango HSI data. Without subspace transfer, HOSVD and FTD yield comparable classification performance. When the training and testing domains differ, however, the results diverge: HOSVD fails to achieve accuracy above random chance, whereas FTD successfully extracts discriminative features, reaching a classification accuracy of $78.95\%$ ($k{=}7$.}
\label{fig:ex3_successrates_lambda_1_d_1}
\end{figure}

Based on the parameters identified in the validation step, the results on the remaining test data are given in terms of the classification accuracy in \Cref{fig:ex3_successrates_lambda_1_d_1} and macro F1 score in \Cref{app:fig:ex3_macrof1scores_lambda_1_d_1}, respectively. As before, the left-side of the charts considers the equal domain in training as in testing, while the right hand-side investigates the subspace transfer for the FTD. 

The left-hand panel of \Cref{fig:ex3_successrates_lambda_1_d_1} illustrates the consistent performance of HOSVD and FTD under identical training conditions. In contrast, the right-hand panel illustrates the effect of the adaptive subspace modeling on the classification results. While the success rates of the HOSVD-based classification do not exceed the baseline of always predicting the majority class RIPE in the classification process, the continuous FTD model achieves a classification accuracy of $78.95\%$ ($k{=}7$). \Cref{app:fig:ex3_macrof1scores_lambda_1_d_1} in the appendix supports this finding by considering the macro F1 score, which compensates for class imbalance by weighting all classes equally. Consequently, the performance gap between HOSVD and FTD becomes even more pronounced in this figure.

\Cref{tab:results_hsi_mango_data} lists the results in comparison to other vector- and tensor-based models. In the same-domain setting, the HOSVD and the FTD achieve the highest accuracy and macro F1 scores ($78.95\%$ and $80.11\%$ for $k{=}7$, respectively). Under subspace transfer the landscape changes: $k$NN collapses to majority-class performance ($42.11\%$ accuracy, macro F1 of $19.75\%$), consistent with the MLP and SVM approaches. The HOSVD likewise degrades to this level, as it has no mechanism to account for the shift in sampling points. The FTD is the only method that retains discriminative power, achieving $78.95\%$ accuracy and a macro F1 score of $80.11\%$ ($k{=}7)$ -- outperforming all non-oracle methods and reaching the oracle ceiling of $78.95\%$ accuracy.
\begin{table}[tbhp]
    \centering
    \caption{\parindent=0pt Classification results on the HSI mango dataset. Training domain: indices \texttt{1:4:224}. Same-domain test: indices \texttt{1:4:224}. Transferred domain test: indices \texttt{100:1:155}. Best results are bold.}
    \label{tab:results_hsi_mango_data}
    \begin{threeparttable}                 
    \small
    \setlength{\tabcolsep}{5pt}    
    \begin{tabular}{@{}lcccc@{}}
        \toprule
        & \multicolumn{2}{c}{Same domain}
        & \multicolumn{2}{c}{Transferred domain} \\
        \cmidrule(lr){2-3}\cmidrule(lr){4-5}
        Model
          & Accuracy (\%)
          & Macro F1 (\%)
          & Accuracy (\%)
          & Macro F1 (\%) \\
        \midrule
        \rowcolor{rowgray}
        \multicolumn{5}{@{}l}{%
          \textit{Supervised baselines (vectorised input)}} \\
        $k$NN ($k{=}1$)\tnote{$\ddagger$}
          & $73.68$ & $73.85$
          & $42.11$ & $19.75$ \\
        MLP\tnote{$*$}
          & $37.37 \pm 0.06$ & $18.04 \pm 0.02$
          & $37.37 \pm 0.06$ & $18.04 \pm 0.02$ \\
        SVM\tnote{$\dagger$}
          & $42.11$ & $19.75$
          & $42.11$ & $19.75$ \\
        \rowcolor{rowgray}
        \multicolumn{5}{@{}l}{%
          \textit{Tensor subspace baseline}} \\
        HOSVD ($k{=}7$)\tnote{$^\circ$}
          & $\bm{78.95}$ & $\bm{80.11}$
          & $42.11$ & $19.75$ \\
        HOSVD ($k{=}9$)\tnote{$^\circ$}
          & $\bm{78.95}$ & $\bm{80.11}$
          & $42.11$ & $19.75$ \\
        \rowcolor{rowgray}
        \multicolumn{5}{@{}l}{%
          \textit{Our method}} \\
        FTD ($k{=}7$)\tnote{$\S$}
          & $\bm{78.95}$ & $\bm{80.11}$
          & $\bm{78.95}$ & $\bm{80.11}$\\
        FTD ($k{=}9$)\tnote{$\S$}
          & $73.68$ & $73.81$
          & $\bm{78.95}$ & $78.36$\\
        \rowcolor{rowgray}
        \multicolumn{5}{@{}l}{%
          \textit{Oracle models}} \\
        HOSVD-Oracle ($k{=}7$)\tnote{$\P$}
          & \multicolumn{1}{c}{---} & \multicolumn{1}{c}{---}
          & $52.63$ & $51.94$ \\
        HOSVD-Oracle ($k{=}9$)\tnote{$\P$}
          & \multicolumn{1}{c}{---} & \multicolumn{1}{c}{---}
          & $\bm{78.95}$ & $78.36$ \\
        \bottomrule
    \end{tabular}
    \begin{tablenotes}[flushleft]\footnotesize
    \item[$*$] MLP: hidden layers $(50,50)$, tanh activation, $\alpha = 0.0001$, determined by grid search. Applied to vectorized slices from the training domain; no subspace transfer is performed at test time. The results are mean $\pm$ std over 10 independent runs.
    \item[$\dagger$] SVM: RBF kernel, $\gamma = 1/(n_{\mathrm{features}} \cdot \mathrm{var}(X))$. Applied to vectorized slices from the training domain; no subspace transfer is performed at test time.
    \item[$\ddagger$] $k$NN: Euclidean distance on vectorised slices, $k$ selected by 5-fold cross-validation on training data. No subspace transfer is performed at test time.
    \item[$^\circ$] HOSVD: Tucker ranks selected via HOSVD-based 5-fold cross-validation set to $(5,5,14)$ (see \Cref{subsec:exp_hsi}).
    \item[$\S$] FTD: Tucker ranks selected via HOSVD-based 5-fold cross-validation set to $(5,5,14)$; regularization parameter and Gaussian kernel bandwidth set to $\lambda=1$ and $d=1$ (see \Cref{subsec:exp_hsi}).
    \item[$\P$] Oracle: model trained and evaluated entirely within the transfer test domain. Represents the performance under no domain shift. Tucker ranks set equal to $(5,5,14)$.
    \end{tablenotes}
    \end{threeparttable}
\end{table}

\section{Conclusions}\label{sec:conclusions}

In this paper, we proposed a functional Tucker decomposition (FTD) which models continuity in one tensor mode using the flexible representation capabilities of reproducing kernel Hilbert spaces (RKHS). Building on this concept, we developed an adaptive subspace modeling framework that enables subspace transfer in the continuous mode, allowing for applications such as classification.

The proposed approach joins recent advances in functional tensor modeling \cite{timmerman2002three, beylkin2005algorithms, dolgov2021functional, SorberVanBarelDeLathauwer2015, TangKoldaZhang2026, LarsenEtAl2024, HanShiZhang2024, QianEtAl2025}. However, the FTD builds on the Tucker decomposition (TD) instead of the Canonical Polyadic decomposition (CPD). This choice inherently shifts emphasis toward subspace structure rather than explicit interpretation of individual modes. This change of perspective proves helpful when leveraging mode-wise subspace information for the purpose of cross-domain classification, as was shown in three different experiments. Furthermore, from a technical perspective, the model setup driven by RKHS provides a broad and rigorous framework with computational tractability, structural flexibility and smoothness control.

The theory was validated with an expanded version of the digit classification problem and demonstrated on multi-dimensional, real-world datasets, including trip-record multivariate time-series and hyperspectral imaging data. The application to 2024 NYC taxi trip records stands as an example for the broad range of time series available across various domains such as economics, healthcare and climate sciences. As the assumptions of continuity generally hold for time-series, our method has the potential to enrich multi-way methods such as classification in most of these areas. The analysis of hyperspectral images (HSIs) presented a rich playing field for our framework due to the multi-dimensional structure of hyperspectral images and their assumed continuity in the spectral mode. Here, the move towards classification combined the perks of the FTD: equivalence of subspace information and adaptive subspace modeling. As HSI sampling is time-consuming and expensive, the FTD framework might offer remedy, as adaptive subspace modeling enables smooth transitions between different camera setups while still keeping the data compatible with solid tensor-based approaches. 

While the proposed adaptive subspace modeling with the FTD is effective and proves beneficial in real-world applications, expansions and further theoretical developments are possible. For instance, an important open research question is the design point selection. Although this selection may be application‑dependent or constrained by measurement availability, investigating general strategies for determining optimal sampling points in RKHS models could yield more stable and adaptable methods. Such strategies may even merit consideration during the planning of data acquisition. Further extensions of the present method include adaptations that target missing data or unaligned sampling schemes, in line with \cite{LarsenEtAl2024,TangKoldaZhang2026}. In addition, transferring the continuity constraint to alternative tensor decompositions -- such as the tensor train decomposition -- presents a promising direction for future work.

\bibliographystyle{plain}
\bibliography{references}

\section*{Acknowledgments}
This work was supported by KU Leuven under project C2E/23/007 and by the Research Foundation – Flanders (FWO) through the fundamental research fellowship 11A2H25N.


\appendix\label{sec:appendix}

\section{Classification using HOSVD}\label{app:subsec:class_hosvd}
The classification method considered in this work is a multidimensional variant of the \textit{Soft Independent Modeling by Class Analogy} (SIMCA) algorithm \cite{Wold1976}, in its high-dimensional version initially presented by Savas and Eldén \cite{SavasElden2007}. The approach follows the principle of class modeling (CM) by constructing a small set of basis arrays from the HOSVD of the tensor of stacked samples for each class with the aim of describing dominant properties. The algorithm operates in two steps: first, within the training stage, class models are constructed from the training data by truncation of the HOSVD of the tensor of stacked samples. Second, in the testing stage, a residual between an input with unknown class and each class model is computed and used to propose a class. Even though the seminal paper of this method did not consider the optimization of model parameters or the validation of the model in separate steps, it is noted that a full implementation of class modeling by means of SIMCA includes both these important steps. Further information on model parameter optimization and model validation within the SIMCA regime can be found in \cite{VitaleEtAl2023}.
\paragraph{Training stage}
Let $C$ denote the number of classes. Without loss of generality, assume that each sample is of the form $\textbf{X}\in\mathbb{R}^{n \times p}$. The first step of the method works on training data which consists of all training samples of a class $\mu\in[C]$ stacked along the first mode, yielding the tensor $\mathbfcal{X}_{\text{training}}^{\mu} \in \mathbb{R}^{M_\mu \times n \times p}$, where $M_\mu$ is the number of training samples for class $\mu$ and $n \times p$ is the size of each training sample. A rank-($M_\mu$,$r$,$s$) Tucker decomposition of $\mathbfcal{X}_{\text{training}}^{\mu}$ is computed via the HOSVD as
\begin{align*}
    \mathbfcal{X}_{\text{training}}^{\mu}~\approx~\mathbfcal{G}^{\mu} \bullet_1 \textbf{A}^{\mu} \bullet_2 \textbf{B}^{\mu} \bullet_3 \textbf{C}^{\mu}
\end{align*}
for $r\leq n$ and $s\leq p$. Then, by setting
\begin{align*}
    \textbf{D}^{\mu} = \mathbfcal{G}^{\mu} \bullet_2 \textbf{B}^{\mu} \bullet_3 \textbf{C}^{\mu} 
    \ \text{ and } \ 
    \textbf{D}_{\nu}^{\mu} = \textbf{D}^{\mu}(\nu,:,:)
\end{align*}
for all $\nu \in [M_\mu]$, it holds 
\begin{align*}
    \mathbfcal{X}_{\text{training}}^{\mu}~\approx~\textbf{D}^{\mu} \bullet_1 \textbf{A}^{\mu}
\end{align*}
and the matrices $\{ \textbf{D}_{1}^{\mu},\ldots,\textbf{D}_{k}^{\mu}\}$ span a small and dominant subspace that characterizes class $\mu$ for a truncation parameter $k<M_\mu$. In other words, the basis for class $\mu$ is given by
\begin{align*}
    \textbf{T}^{\mu} = \left( \textbf{D}_{\nu}^{\mu} \right)_{\nu = 1}^{k}
\end{align*}
where each $\textbf{D}_{\nu}^{\mu}$ is normalized. Here, we note that the computation via the HOSVD guarantees orthogonality of $\textbf{D}_{\nu}^{\mu}$ for all $\nu$ and $\mu$ due to the all-orthogonality of the core $\mathbfcal{G}^{\mu}$; this will prove helpful in the test stage. Finally, this process is repeated for all classes $\mu_1,\ldots,\mu_C$, yielding the corresponding class-wise bases $\textbf{T}^{\mu_1},\ldots,\textbf{T}^{\mu_C}$. 

\paragraph{Test stage}
Let $\textbf{Y}\in\mathbb{R}^{n \times p}$ be a normalized test sample without a class label. In order to assign it to a class, the goal is to find the basis which provides the best approximation of $\textbf{Y}$. This translates to the minimization problems
\begin{align*}
    \min_{\alpha_{1}^{\mu},\ldots,\alpha_{k}^{\mu}} || \textbf{Y} - \sum_{\nu=1}^{k} \alpha_{\nu}^{\mu} \textbf{D}_{\nu}^{\mu} ||
\end{align*}
with unknowns $\alpha_{1}^{\mu},\ldots,\alpha_{k}^{\mu}$. Owing to the orthonormality of $\textbf{D}_{\nu}^{\mu}$ the solution is given by $\widehat{\alpha}_\nu^\mu = \langle \textbf{Y}, \textbf{D}_{\nu}^{\mu} \rangle.$ As a result, the residual can be computed via
\begin{align*}
    R(\mu) 
    &= || \textbf{Y} - \sum_{\nu=1}^{k} \widehat{\alpha}_{\nu}^{\mu} \textbf{D}_{\nu}^{\mu} ||^2  = 1 - \sum_{\nu=1}^{k} \langle \textbf{Y}, \textbf{D}_\nu^\mu \rangle^2
\end{align*}
according to the orthonormality of $\textbf{D}_{\nu}^{\mu}$ for all $\nu\in[k]$ \cite{SavasElden2007}. At last, the test sample $\textbf{Y}\in\mathbb{R}^{n \times p}$ is assigned to the class $\mu^*$ where $R(\mu^*)$ is minimal.

\section{Reproducing kernel Hilbert spaces (RKHS)}\label{app:subsec:rkhs}

For the purpose of modeling the functional mode in the Tucker decomposition, the functions are chosen from a \textit{reproducing kernel Hilbert space} (RKHS). A RKHS is a Hilbert space of functions with useful representation properties through the notion of a kernel. More precisely, for a non-empty set $I=\{x_1,\ldots,x_p\}$ with $x_i\in\mathbb{R}$ for all $i\in[p]$, let ${\mathcal{K}:I \times I \rightarrow \mathbb{R}}$ be a symmetric positive semidefinite (psd) kernel function with corresponding kernel matrix $\textbf{K}$ defined by its elements $\mathcal{K}_{ij} = \mathcal{K}(x_i,x_j)$. A prominent example is the \textit{Gaussian kernel} which is defined via 
\begin{align*}
    \mathcal{K}(x_i,x_j) = \exp\left(\frac{-(x_i-x_j)^2}{2d^2}\right)
\end{align*}
for a parameter $d\neq0$. In the following, functions $f$ and $g$ are constructed via
\begin{align*}
    f(\cdot) = \sum_{i=1}^{p} \alpha_i \mathcal{K}(\cdot,x_i) \quad \text{and} \quad g(\cdot) = \sum_{j=1}^{p} \beta_j \mathcal{K}(\cdot,x_j)
\end{align*}
for $\alpha_i,\beta_j\in\mathbb{R}$ and $x_i,x_j\in I$ for $i,j\in[p]$. This leads to the natural definition of a dot product by $\langle f,g \rangle_\mathcal{K} = \sum_{i,j=1}^{p} \alpha_i \beta_j \mathcal{K}(x_i,x_j)$ and, accordingly, the induced norm $||f||_\mathcal{K}:=\sqrt{\langle f,f \rangle_\mathcal{K}}$. Now, the space of linear combinations of the kernel which is completed with respect to $||f||_\mathcal{K}$, i.e., 
\begin{align*}
    \mathcal{H}_\mathcal{K} := \overline{\linspan \{ \mathcal{K}(\cdot,x_i) \ | \ x_i \in I\}},
\end{align*}
is called a reproducing kernel Hilbert space (RKHS). The important \textit{reproducing kernel property} states that
\begin{align*}
    \langle \mathcal{K}(\cdot,x_i), f \rangle_\mathcal{K} = f(x_i)
\end{align*}
for $x_i\in I$. This means that a function $f$ at a point $x_i$ can be evaluated by taking an inner product with the evaluated kernel $\mathcal{K}(\cdot,x_i)$ \cite{SchoelkopfSmola2001}. Furthermore, \Cref{thm:repr_thm} displays the \textit{Representer theorem} which states that a large subset of infinite dimensional optimization problems over $\mathcal{H}_\mathcal{K}$ admits solutions in terms of linear combinations of the kernel evaluated on a finite set of particular design points $\{\mathcal{K}(\cdot,x_1),\ldots,\mathcal{K}(\cdot,x_p)\}$.
\begin{theorem}[Representer theorem \cite{SchoelkopfSmola2001}]\label{thm:repr_thm}
    Denote by $\Omega:[0,\infty) \rightarrow \mathbb{R}$ a strictly monotonic increasing function, by $I=\{x_1,\ldots,x_p\}$ a set, and by $c:(I \times \mathbb{R}^2)^p \rightarrow \mathbb{R} \cup \{ \infty \}$ an arbitrary loss function. Then each minimizer $f^*\in\mathcal{H}_\mathcal{K}$ of the regularized risk
    \begin{align*}
        c\left( (x_1,y_1,f^*(x_1)),\ldots,(x_p,y_p,f^*(x_p)) \right) + \Omega(||f^*||_\mathcal{K})
    \end{align*}
    admits a representation of the form
    \begin{align*}
        f^*(\cdot) = \sum_{i=1}^{p} w_i \mathcal{K}(\cdot,x_i).
    \end{align*}
\end{theorem}
In the following, analogous to \cite{TangKoldaZhang2026}, we only consider the important variant of \Cref{thm:repr_thm}, originally stated by \cite{KimeldorfWahba1971}, focusing on the point-wise mean squared loss
\begin{equation*}
    c( (x_1,y_1,f(x_1)), \ldots, (x_p,y_p,f(x_p)) ) = \frac{1}{p} \sum_{i=1}^{p} (y_i - f(x_i))^2
\end{equation*}
and $\Omega(||f||_\mathcal{K}) = \frac{\lambda}{2}||f||_\mathcal{K}^2$ for $\lambda > 0$. This setup guarantees an accessible solution of the form 
\begin{align*}
    f^*(\cdot) = \sum_{i=1}^{p} w_i \mathcal{K}(\cdot,x_i) = \widetilde{\mathbf{K}} \textbf{w},
\end{align*}
for some $\textbf{w}\in\mathbb{R}^p$ and a quasimatrix $\widetilde{\mathbf{K}} = \mathcal{K}(\cdot,\textbf{x}) \in\mathbb{R}^{\infty \times p}$ for $\textbf{x}=\trans{[x_1 \ x_2 \ldots x_p]}$ which -- evaluated at the design points in $I$ -- corresponds to the kernel matrix $\textbf{K} \in\mathbb{R}^{p \times p}$. This representation will prove favorable to the optimization problem in \Cref{sec:functional_tucker_decomposition}.

\section{Complexity considerations}\label{app:subsec:complexity}

In each iteration of \Cref{alg:ftd}, the factor matrices $\textbf{A}$, $\textbf{B}$ and $\textbf{W}$ are computed independently. 

\paragraph{Discrete factor matrices} For the discrete factor matrix $\textbf{A}$, computing the Kronecker product of $\textbf{C}$ and $\textbf{B}$ costs $\mathcal{O}(psnr)$ floating-point operations (flops). Adding the costs of the matrix multiplication between $\mathbf{G}_{1}$ and $\trans{(\textbf{C}\otimes\textbf{B})}$ and the solution of the linear system with the QR decomposition, amounting to $\mathcal{O}(qrspn)$ and $\mathcal{O}(pnq^2)$ flops, respectively, results in costs of $\mathcal{O}(q^2pn + qrspn)$ flops. Moreover, computing the SVD of $\textbf{A}_{1}$ takes $\mathcal{O}(mq^2)$ flops, while the $1$-mode product $\mathbfcal{G}\bullet_1(\Sigma_1\trans{\textbf{V}_{1}})$ costs $\mathcal{O}(q^2rs)$ flops. In total, this adds up to $\mathcal{O}(q^2(m+np) + qrspn)$ flops for the computation of $\textbf{A}$ and the following update of $\mathbfcal{G}$. Exploiting the symmetry between the computation of $\textbf{A}$ and $\textbf{B}$, results in a complexity of $\mathcal{O}(r^2(n+mp)+qrspm)$ flops for the update of $\textbf{B}$ and $\mathbfcal{G}$ in the second block of the algorithm. 

\paragraph{Continuous factor matrix and core tensor} Turning to the continuous factor matrix $\textbf{C}$, the computation of the placeholder matrices $\textbf{L}$ and $\textbf{R}$ costs $\mathcal{O}(s^2(qr+p^2))$ and $\mathcal{O}(prq(mn+s))$ flops, respectively. The following solution of the linear system with Cholesky accounts to $\mathcal{O}((ps)^3)$ flops. Moreover, the factor matrix $\textbf{C}$ is updated with $\mathcal{O}(p^2s)$ flops, before the final projection of the core tensor takes $\mathcal{O}(mnp(q+r) + ps^2))$ flops.

In total, the final block of the algorithm requires $\mathcal{O}(p^3s^3 + qr(s^2 + pmn))$ flops for the computation of the continuous factor matrix and $\mathcal{O}(mnp(q+r) + ps^2))$ for the update of the core tensor. Overall, each iteration of \Cref{alg:ftd} comprises $\mathcal{O}(p^3s^3 + q^2m + r^2n + qrspn + qrspm + qrpmn)$ flops.

\section{Reconstruction error under subspace transfer}\label{app:subsec:reconstr_err}

The following derivation of the interpolation-error bound via the (unregularized) power function adapts the classical
scattered-data approximation argument of \cite{Wendland2004}, in which the power function measures the pointwise error of the minimum-norm interpolant. The underlying RKHS machinery follows \cite{SchoelkopfSmola2001}. Our regularized power function $P_{\mathcal{K},\lambda}$ extends this classical interpolation analysis to the regularized smoother by replacing $\mathbf{K}^{-1}$ with $(\mathbf{K}+\lambda\mathbf{I}_p)^{-1}$, which is the modification specific to the FTD estimator.

\subsection{Setup}\label{app:subsubsec:reconstr_err_setup}

Write $\mathcal{K}(\cdot,\cdot)$ for the kernel function and $\textbf{K}=[\mathcal{K}(x_i,x_j)]_{i,j=1}^{p}\in\mathbb{R}^{p\times p}$ for the kernel matrix on the design set $I=\{x_1,\ldots,x_p\}\subset\mathbb{R}$. For a point $x\in\mathbb{R}$ set
\begin{align*}
    \textbf{k}(\cdot) = \trans{(\mathcal{K}(\cdot,x_1),\ldots,\mathcal{K}(\cdot,x_p))},
    \qquad
    \textbf{k}(x) = \trans{(\mathcal{K}(x,x_1),\ldots,\mathcal{K}(x,x_p))}\in\mathbb{R}^{p}.
\end{align*}
Let $\mathcal{H}_{\mathcal{K}}$ be the RKHS of $\mathcal{K}$ with inner product
$\langle\cdot,\cdot\rangle_{\mathcal{H}_{\mathcal{K}}}$, let
$V_I:=\operatorname{span}\{\mathcal{K}(\cdot,x_i):x_i\in I\}\subset\mathcal{H}_{\mathcal{K}}$,
and for $g\in\mathcal{H}_{\mathcal{K}}$ write $g(I)=\trans{(g(x_1),\ldots,g(x_p))}$.

The continuous factor of the FTD is obtained from
\begin{align}
    \min_{\mathbfcal{G},\textbf{A},\textbf{B},\textbf{W}}
    \;\tfrac{1}{2}\,\big\|\,\widehat{\mathbfcal{T}}_{I}
      - \mathbfcal{G}\bullet_1\textbf{A}\bullet_2\textbf{B}\bullet_3\textbf{K}\textbf{W}\,\big\|^2
      + \frac{\lambda_{\mathrm{F}}}{2}\,\|\textbf{W}\|_{\textbf{K}}^2,\label{app:eq:ftd_optimization}
\end{align}
with FTD regularization parameter $\lambda_{\mathrm{F}}>0$.
Given the optimal weights $(\textbf{W}=[\mathbf{w}_1\cdots\mathbf{w}_s]\in\mathbb{R}^{p\times s})$,
the $\sigma$-th column of the functional factor is represented in $\mathcal{H}_{\mathcal{K}}$ as
\begin{align*}
    \hat{c}_{\sigma}(\cdot)
    = \sum_{h=1}^{p} w_{h\sigma}\,\mathcal{K}(\cdot,x_h)
    = \trans{\textbf{k}(\cdot)}\,\mathbf{w}_\sigma,
    \qquad \sigma\in[s].
\end{align*}

Subspace transfer evaluates the learned factor on a second, in general distinct, transfer domain $\tilde{I}=\{\tilde{x}_1,\ldots,\tilde{x}_{|\tilde I|}\}\subset\mathbb{R}$.
Collecting the true and estimated functional columns sampled on $\tilde I$ into
\begin{align*}
    \mathbf{C}^{*}_{\tilde I}=[c^{*}_1(\tilde I)\;\cdots\;c^{*}_s(\tilde I)],
    \qquad
    \hat{\mathbf{C}}_{\tilde I}=[\hat{c}_1(\tilde I)\;\cdots\;\hat{c}_s(\tilde I)],
\end{align*}
the quantity of interest is the reconstruction error in the transfer domain,
\begin{align}
    \big\|\hat{\mathbf{C}}_{\tilde I}-\mathbf{C}^{*}_{\tilde I}\big\|.
    \label{eq:reconstr_target}
\end{align}
We control \eqref{eq:reconstr_target} column-by-column through the pointwise RKHS error
\begin{align}
    \big|c_{\sigma}^{*}(\tilde x)-\hat{c}_{\sigma}(\tilde x)\big|,
    \qquad \tilde x\in\tilde I,\ \sigma\in[s],
    \label{eq:rkhs_error}
\end{align}
where $c_{\sigma}^{*}\in\mathcal{H}_{\mathcal{K}}$ is the true $\sigma$-th functional column.
The argument proceeds in three steps.
\begin{enumerate}[label=(\roman*), ref=\roman*] 
    \item The subproblem for the continuous mode in (\ref{app:eq:ftd_optimization}) is a generalized Sylvester equation. Under assumption~\ref{ass:core_unf_full_rank}, an orthogonal change of variables decouples it into $s$ independent kernel ridge regressions with effective regularization parameters $\lambda_{\mathrm{F}}/\mu_\sigma$, reducing the analysis to a single ridge regression.
    \item The pointwise error \eqref{eq:rkhs_error} of one such ridge regression is bounded using the minimum-norm interpolant as a tractable intermediary, which yields a closed form for the (regularized) power function.
    \item The column-wise bounds are reassembled into a bound on \eqref{eq:reconstr_target}; orthogonality of the change of variables leaves both the Frobenius error and the aggregate RKHS norm $\|\mathbf{C}^{*}\|_{\mathcal{H}_\mathcal{K}}$ invariant.
\end{enumerate}

\subsection{Decoupling the system}\label{app:subsubsec:reconstr_err_decoupling}

We first argue why the continuous-factor update solves a coupled system, and how assumptions~\ref{ass:core_unf_full_rank} and \ref{ass:A_B_orth} reduce it, without loss of generality, to a single kernel ridge regression. According to \Cref{par:solution_continuous_mode}, the inner update for the continuous factor in \eqref{eq:ftd_optimization_raw} fixes $\mathbfcal{G},\textbf{A},\textbf{B}$ and solves
\begin{align*}
    \min_{\textbf{W}}
    \tfrac{1}{2}\,\big\|\,\widehat{\mathbfcal{T}}_{I}
      - \mathbfcal{G}\bullet_1\textbf{A}\bullet_2\textbf{B}\bullet_3\textbf{K}\textbf{W}\,\big\|^2
      + \frac{\lambda_{\mathrm{F}}}{2}\,\|\textbf{W}\|_{\textbf{K}}^2 .
\end{align*}
With $\textbf{M}=\textbf{G}_{(3)}\trans{\textbf{G}}_{(3)}$ as in
assumption~\ref{ass:core_unf_full_rank} and $\textbf{A},\textbf{B}$ orthonormal
(assumption~\ref{ass:A_B_orth}), the associated normal equations are equivalent to the
generalized Sylvester equation
\begin{align*}
    \textbf{K}\textbf{W}\textbf{M} + \lambda_{\mathrm{F}}\,\textbf{W} = \textbf{Y},
    \qquad
    \textbf{Y} := \widehat{\textbf{T}}_{(3)}(\textbf{B}\otimes\textbf{A})\trans{\textbf{G}}_{(3)} .
\end{align*}
Since $\textbf{G}_{(3)}$ is not orthogonal, $\textbf{M}\neq\textbf{I}_s$, and $\textbf{M}$ acts
as a coupling operator across the columns of $\textbf{W}$,
\begin{align*}
    \sum_{j=1}^{s}\textbf{M}_{ji}\,\textbf{K}\textbf{w}_{j} + \lambda_{\mathrm{F}}\,\textbf{w}_{i}
    = \textbf{Y}_{:,i},
    \qquad i\in[s],
\end{align*}
which obstructs a direct column-wise error bound.

By assumption~\ref{ass:core_unf_full_rank}, $\textbf{M}$ is symmetric positive definite and
admits a spectral decomposition $\textbf{M}=\textbf{Q}\,\Lambda\,\trans{\textbf{Q}}$ with
$\Lambda=\operatorname{diag}(\mu_1,\dots,\mu_s)$, $0<\mu_{1}\le\cdots\le\mu_{s}$, and
$\trans{\textbf{Q}}\textbf{Q}=\textbf{I}_s$. Right-multiplying the Sylvester equation by
$\textbf{Q}$ and using $\textbf{M}\textbf{Q}=\textbf{Q}\Lambda$ uncouples it: with
$\textbf{W}'=\textbf{W}\textbf{Q}$ and $\textbf{Y}'=\textbf{Y}\textbf{Q}$,
\begin{align*}
    \textbf{K}\textbf{W}'\Lambda + \lambda_{\mathrm{F}}\,\textbf{W}' = \textbf{Y}',
    \qquad\text{i.e.}\qquad
    (\mu_\sigma\textbf{K} + \lambda_{\mathrm{F}}\textbf{I}_p)\,\textbf{w}'_\sigma = \textbf{y}'_\sigma,
    \quad \sigma\in[s].
\end{align*}

The change of variables $\textbf{W}'=\textbf{W}\textbf{Q}$ is orthogonal, so it leaves the two quantities entering the final bound invariant: writing
$\hat{\textbf{C}}'_{\tilde I}=\hat{\textbf{C}}_{\tilde I}\textbf{Q}$ and $\textbf{C}'^{*}_{\tilde I}=\textbf{C}^{*}_{\tilde I}\textbf{Q}$,
\begin{align*}
    \big\|\hat{\textbf{C}}_{\tilde I}-\textbf{C}^{*}_{\tilde I}\big\|
    = \big\|\hat{\textbf{C}}'_{\tilde I}-\textbf{C}'^{*}_{\tilde I}\big\|,
    \qquad
    \sum_{\sigma=1}^{s}\big\|c'^{*}_\sigma\big\|_{\mathcal{H}_{\mathcal{K}}}^2
    = \sum_{\sigma=1}^{s}\big\|c^{*}_\sigma\big\|_{\mathcal{H}_{\mathcal{K}}}^2
    = \|\textbf{C}^{*}\|_{\mathcal{H}_\mathcal{K}}^2 ,
\end{align*}
the second identity following from $\trans{\textbf{Q}}\textbf{Q}=\textbf{I}_s$ and invariance of
the trace of the Gram matrix of $\{c^{*}_\sigma\}$. It therefore suffices to bound a single
column. Dropping the prime, we write $c^{*}_\sigma,\hat{c}_\sigma$ for a generic
rotated column pair and bound the well-specified ridge regression
$\hat{c}_\sigma=S^{\lambda}_{c^{*}_\sigma}$ for a generic parameter $\lambda>0$, which for
column $\sigma$ equals $\lambda_{\mathrm{F}}/\mu_\sigma$.

\subsection{Minimum norm interpolant}\label{app:subsubsec:reconstr_err_minnorminterp}

By the reduction of \Cref{app:subsubsec:reconstr_err_decoupling}, it suffices to bound a single
well-specified ridge regression $\hat{c}_\sigma=S^{\lambda}_{c^{*}_\sigma}$ with generic parameter
$\lambda>0$. As a tractable intermediary, we first treat the unregularized ($\lambda=0$) case and
derive the minimum-norm interpolant $S_g$ of an arbitrary function $g\in\mathcal{H}_{\mathcal{K}}$ that
minimizes
\begin{align}
    \min_{f\in\mathcal{H}_{\mathcal{K}}} \|f\|_{\mathcal{H}_{\mathcal{K}}}
    \quad\text{subject to}\quad f(x_i)=g(x_i)\ \text{for all } x_i\in I.
    \label{eq:minnorminterp_opt}
\end{align}
The interpolant $S_{c^{*}_\sigma}$ provides the power-function machinery used to bound
\Cref{eq:rkhs_error}; its regularized analogue, applicable to the FTD estimator $\hat{c}_\sigma$
itself, is derived in the next subsection.
 
\begin{lemma}\label[lemma]{lem:minnorminterp}
    Under assumption~\ref{ass:kernel_matr_pd}, the minimum-norm interpolant $S_g$ of a target
    $g\in\mathcal{H}_{\mathcal{K}}$ that minimizes \eqref{eq:minnorminterp_opt} is
    \begin{align*}
        S_{g}(\cdot) = \trans{\textbf{k}(\cdot)}\,\textbf{K}^{-1}g(I).
    \end{align*}
\end{lemma}
\begin{proof}
Consider a candidate $f(\cdot) = \sum_{i=1}^{p}\alpha_i \mathcal{K}(\cdot, x_i) + f^\perp(\cdot)$,
where $f^\perp \perp V_I$. The interpolation conditions give
\[
    f(x_j) = \sum_{i=1}^{p}\alpha_i \mathcal{K}(x_j, x_i) + f^\perp(x_j) = g(x_j)
    \quad \text{for all } x_j \in I.
\]
Since $f^\perp \perp V_I$, the reproducing property yields
$f^\perp(x_j) = \langle f^\perp, \mathcal{K}(\cdot, x_j)\rangle_{\mathcal{H}_{\mathcal{K}}} = 0$ for all
$x_j\in I$, so the conditions reduce to $\textbf{K}\,\boldsymbol{\alpha} = g(I)$. As $\textbf{K}$ is
positive definite (assumption~\ref{ass:kernel_matr_pd}), it is invertible and
$\boldsymbol{\alpha} = \textbf{K}^{-1}g(I)$. By orthogonality,
\begin{align*}
    \|f\|^2_{\mathcal{H}_{\mathcal{K}}}
    = \Bigl\|\sum_i \alpha_i \mathcal{K}(\cdot,x_i)\Bigr\|^2_{\mathcal{H}_{\mathcal{K}}}
      + \|f^\perp\|^2_{\mathcal{H}_{\mathcal{K}}}
    \;\geq\;\Bigl\|\sum_i \alpha_i \mathcal{K}(\cdot,x_i)\Bigr\|^2_{\mathcal{H}_{\mathcal{K}}},
\end{align*}
with equality if and only if $\|f^\perp\|^2_{\mathcal{H}_{\mathcal{K}}} = 0$, i.e.\ $f^\perp = 0$.
Hence the unique minimum is $S_{g}(\cdot) = \trans{\textbf{k}(\cdot)}\,\textbf{K}^{-1}g(I)$.
\end{proof}

\subsection{Bounding the interpolation error}\label{app:subsubsec:reconstr_err_boundinginterperr}
 
With the minimum-norm interpolant $S_{c_{\sigma}^{*}}$, we bound
$|c_{\sigma}^{*}(\tilde{x}) - S_{c_{\sigma}^{*}}(\tilde{x})|$ for a transfer point
$\tilde{x}\in\tilde{I}$. By the reproducing property,
\begin{align*}
    c_{\sigma}^{*}(\tilde{x}) = \langle c_{\sigma}^{*},\,\mathcal{K}(\cdot,\tilde{x})\rangle_{\mathcal{H}_{\mathcal{K}}},
    \qquad
    S_{c_{\sigma}^{*}}(\tilde{x}) = \langle S_{c_{\sigma}^{*}},\,\mathcal{K}(\cdot,\tilde{x})\rangle_{\mathcal{H}_{\mathcal{K}}},
\end{align*}
so that
\begin{align*}
    c_{\sigma}^{*}(\tilde{x}) - S_{c_{\sigma}^{*}}(\tilde{x})
    = \langle c_{\sigma}^{*} - S_{c_{\sigma}^{*}},\;\mathcal{K}(\cdot,\tilde{x})\rangle_{\mathcal{H}_{\mathcal{K}}}.
\end{align*}
Decomposing $\mathcal{K}(\cdot,\tilde{x})$ orthogonally as
\begin{align*}
    \mathcal{K}(\cdot,\tilde{x}) = S_{\mathcal{K}(\cdot,\tilde{x})}
      + \bigl(\mathcal{K}(\cdot,\tilde{x}) - S_{\mathcal{K}(\cdot,\tilde{x})}\bigr),
\end{align*}
where $S_{\mathcal{K}(\cdot,\tilde{x})}(\cdot)=\trans{\textbf{k}(\cdot)}\textbf{K}^{-1}\textbf{k}(\tilde{x})
=\sum_{i=1}^{p}\alpha_i\,\mathcal{K}(\cdot,x_i)$, with
$\boldsymbol{\alpha}:=\textbf{K}^{-1}\textbf{k}(\tilde{x})$, is the minimum-norm interpolant of the
kernel section $\mathcal{K}(\cdot,\tilde{x})$ at $I$ (\Cref{lem:minnorminterp} with
$g=\mathcal{K}(\cdot,\tilde{x})$, $g(I)=\textbf{k}(\tilde{x})$), we obtain
\begin{align*}
    c_{\sigma}^{*}(\tilde{x}) - S_{c_{\sigma}^{*}}(\tilde{x})
    = \langle c_{\sigma}^{*} - S_{c_{\sigma}^{*}},\;
      \mathcal{K}(\cdot,\tilde{x}) - S_{\mathcal{K}(\cdot,\tilde{x})}\rangle_{\mathcal{H}_{\mathcal{K}}},
\end{align*}
since $c_{\sigma}^{*} - S_{c_{\sigma}^{*}} \perp V_I$ and $S_{\mathcal{K}(\cdot,\tilde{x})} \in V_I$. By the
Cauchy--Schwarz inequality,
\begin{align*}
    |c_{\sigma}^{*}(\tilde{x}) - S_{c_{\sigma}^{*}}(\tilde{x})|
    \;\leq\;
    \|c_{\sigma}^{*} - S_{c_{\sigma}^{*}}\|_{\mathcal{H}_{\mathcal{K}}}\;\cdot\;
    \|\mathcal{K}(\cdot,\tilde{x}) - S_{\mathcal{K}(\cdot,\tilde{x})}\|_{\mathcal{H}_{\mathcal{K}}}.
\end{align*}
The first factor is bounded via
$\|c_{\sigma}^{*} - S_{c_{\sigma}^{*}}\|_{\mathcal{H}_{\mathcal{K}}} \leq \|c_{\sigma}^{*}\|_{\mathcal{H}_{\mathcal{K}}}$,
as $S_{c_{\sigma}^{*}}$ is an orthogonal projection and does not increase norms. The second factor,
\begin{align*}
    \|\mathcal{K}(\cdot,\tilde{x}) - S_{\mathcal{K}(\cdot,\tilde{x})}\|_{\mathcal{H}_{\mathcal{K}}}
    =: P_{\mathcal{K},0}(\tilde{x},I),
\end{align*}
is the (unregularized) power function, addressed next.


\subsubsection*{Closed form for the power function}\label{app:subsubsec:reconstr_err_closedform_power}
 
Expanding $P_{\mathcal{K},0}(\tilde{x},I)^2$ yields three summands,
\begin{align*}
    P_{\mathcal{K},0}(\tilde{x},I)^2
    = \|\mathcal{K}(\cdot,\tilde{x}) - S_{\mathcal{K}(\cdot,\tilde{x})}\|^2_{\mathcal{H}_{\mathcal{K}}}
    &= \underbrace{\|\mathcal{K}(\cdot,\tilde{x})\|^2_{\mathcal{H}_{\mathcal{K}}}}_{(\mathrm{I})}
      - 2\underbrace{\langle \mathcal{K}(\cdot,\tilde{x}),\,S_{\mathcal{K}(\cdot,\tilde{x})}\rangle_{\mathcal{H}_{\mathcal{K}}}}_{(\mathrm{II})} + \\
      &+ \underbrace{\|S_{\mathcal{K}(\cdot,\tilde{x})}\|^2_{\mathcal{H}_{\mathcal{K}}}}_{(\mathrm{III})},
\end{align*}
which reformulate, using assumption~\ref{ass:kernel_matr_pd}, as
\begin{enumerate}[label=(\Roman*), ref=\Roman*]
\item $\|\mathcal{K}(\cdot,\tilde{x})\|^2_{\mathcal{H}_{\mathcal{K}}}
       = \langle \mathcal{K}(\cdot,\tilde{x}),\,\mathcal{K}(\cdot,\tilde{x})\rangle_{\mathcal{H}_{\mathcal{K}}}
       = \mathcal{K}(\tilde{x},\tilde{x})$;
 
\item $\langle \mathcal{K}(\cdot,\tilde{x}),\,S_{\mathcal{K}(\cdot,\tilde{x})}\rangle_{\mathcal{H}_{\mathcal{K}}}
       = S_{\mathcal{K}(\cdot,\tilde{x})}(\tilde{x})
       = \trans{\textbf{k}(\tilde{x})}\,\textbf{K}^{-1}\textbf{k}(\tilde{x})$;
 
\item $\|S_{\mathcal{K}(\cdot,\tilde{x})}\|^2_{\mathcal{H}_{\mathcal{K}}}
       = \trans{\boldsymbol{\alpha}}\,\textbf{K}\,\boldsymbol{\alpha}
       = \trans{\textbf{k}(\tilde{x})}\,\textbf{K}^{-1}\textbf{K}\,\textbf{K}^{-1}\textbf{k}(\tilde{x})
       = \trans{\textbf{k}(\tilde{x})}\,\textbf{K}^{-1}\textbf{k}(\tilde{x})$.
\end{enumerate}
Combining $(\mathrm{I}) - 2(\mathrm{II}) + (\mathrm{III})$ gives
\begin{align*}
    P_{\mathcal{K},0}(\tilde{x},I)
    = \Bigl(\mathcal{K}(\tilde{x},\tilde{x})
      - \trans{\textbf{k}(\tilde{x})}\,\textbf{K}^{-1}\textbf{k}(\tilde{x})\Bigr)^{1/2}.
\end{align*}
In conclusion, the interpolation error satisfies
\begin{align*}
    |c_{\sigma}^{*}(\tilde{x}) - S_{c_{\sigma}^{*}}(\tilde{x})|
    \;\leq\;
    \|c_{\sigma}^{*}\|_{\mathcal{H}_{\mathcal{K}}}\;
    P_{\mathcal{K},0}(\tilde{x},I)
    \;=\;
    \|c_{\sigma}^{*}\|_{\mathcal{H}_{\mathcal{K}}}
    \Bigl(\mathcal{K}(\tilde{x},\tilde{x})
      - \trans{\textbf{k}(\tilde{x})}\,\textbf{K}^{-1}\textbf{k}(\tilde{x})\Bigr)^{1/2}.
\end{align*}

\subsection{The regularized power function}\label{app:subsubsec:reconstr_err_reg_power}

The interpolation bound above controls the error of the exact interpolant $S_{c_{\sigma}^{*}}$, whose defining property is that it reproduces $c_{\sigma}^{*}$ on the design set $I$. The FTD, however, does not interpolate. Its continuous factor solves the regularized problem with $\lambda>0$, giving $\hat{c}_{\sigma}(\cdot)=\trans{\mathbf{k}(\cdot)}(\mathbf{K}+\lambda\mathbf{I}_p)^{-1}\mathbf{y}_{\sigma}$. Consequently
$\hat{c}_{\sigma}(x_i)\neq c_{\sigma}^{*}(x_i)$ at the design points and the derivation underlying $P_{K,0}$ does not apply to $\hat{c}_{\sigma}$. We therefore replace $P_{K,0}$ by the regularized power function $P_{K,\lambda}$, whose closed form we now derive.\\

The exact interpolant of \Cref{lem:minnorminterp} reproduces $g$ on $I$ and is therefore
unavailable to the FTD, which solves a regularized problem. We replace it by the
\emph{regularized smoother} $S^\lambda$: for $g\in\mathcal{H}_{\mathcal{K}}$, the function $S^\lambda_{g}$ is the
minimizer of the Tikhonov-regularized least-squares problem
\begin{align*}
    \min_{f\in\mathcal{H}_{\mathcal{K}}}\sum_{i=1}^{p}\big(f(x_i)-g(x_i)\big)^2 \;+\; \lambda\,\|f\|_{\mathcal{H}_{\mathcal{K}}}^2 ,
\end{align*}
the regularized analogue of the minimum-norm interpolation problem~\eqref{eq:minnorminterp_opt}
underlying \Cref{lem:minnorminterp}. By the Representer Theorem its minimizer lies in $V_I$
and admits the closed form
\begin{align}\label{eq:Slam-def}
  S^\lambda_{g}(\cdot) \;=\; \trans{\mathbf{k}(\cdot)}\,(\mathbf{K}+\lambda\mathbf{I}_p)^{-1}\,g(I),
\end{align}
which is exactly the interpolant of \Cref{lem:minnorminterp} with $\mathbf{K}^{-1}$ replaced by
$(\mathbf{K}+\lambda\mathbf{I}_p)^{-1}$. Applying \eqref{eq:Slam-def} to the kernel section $\mathcal{K}(\cdot,\tilde{x})$, whose evaluation
vector is $\mathbf{k}(\tilde{x})$, yields the \emph{regularized representer}
\begin{align}\label{eq:reg-representer}
    S^\lambda_{K(\cdot,\tilde{x})}(\cdot)
    &= \trans{\mathbf{k}(\cdot)}\,(\mathbf{K}+\lambda\mathbf{I}_p)^{-1}\,\mathbf{k}(\tilde{x})\\
    &= \sum_{i=1}^{p}\beta_i\,\mathcal{K}(\cdot,x_i)\;\in\;V_I,
  \qquad \boldsymbol{\beta} := (\mathbf{K}+\lambda\mathbf{I}_p)^{-1}\mathbf{k}(\tilde{x}) .
\end{align}
This object obeys a regularized analogue of the reproducing property, which is the identity
that makes the bound below possible.
\begin{lemma}\label[lemma]{lem:reg-reproducing}
For every $g\in\mathcal{H}_{\mathcal{K}}$ and every $\tilde{x}$,
\[
  S^\lambda_{g}(\tilde{x}) \;=\; \big\langle g,\; S^\lambda_{\mathcal{K}(\cdot,\tilde{x})}\big\rangle_{\mathcal{H}_{\mathcal{K}}} .
\]
\end{lemma}
\begin{proof}
Using the representer form \eqref{eq:reg-representer} and the reproducing property
\begin{align*}
    \langle g, K(\cdot,x_i)\rangle_{\mathcal{H}_{\mathcal{K}}} = g(x_i),
\end{align*}
it follows
\begin{align*}
  \big\langle g,\, S^\lambda_{\mathcal{K}(\cdot,\tilde{x})}\big\rangle_{\mathcal{H}_{\mathcal{K}}}
   = \sum_{i=1}^{p}\beta_i\,\langle g, \mathcal{K}(\cdot,x_i)\rangle_{\mathcal{H}_{\mathcal{K}}}
   = \trans{\boldsymbol{\beta}}\,g(I)
   = \trans{\mathbf{k}(\tilde{x})}\,(\mathbf{K}+\lambda\mathbf{I}_p)^{-1}\,g(I)
   = S^\lambda_{g}(\tilde{x}),
\end{align*}
where the third equality uses the symmetry of $(\mathbf{K}+\lambda\mathbf{I}_p)^{-1}$ and the last equality follows from the definition \eqref{eq:Slam-def}.
\end{proof}
We now specialize to the FTD estimator. Its continuous factor is the kernel ridge fit
$\hat{c}_{\sigma}(\cdot)=\trans{\mathbf{k}(\cdot)}(\mathbf{K}+\lambda\mathbf{I}_p)^{-1}\mathbf{y}_{\sigma}$. In the well-specified case
$\mathbf{y}_{\sigma}=c_{\sigma}^{*}:=c_{\sigma}^{*}(I)$, meaning that the targets are the clean samples, comparison with \eqref{eq:Slam-def} gives $\hat{c}_{\sigma} = S^\lambda_{c_{\sigma}^{*}}$
exactly. Combining the reproducing property $c_{\sigma}^{*}(\tilde{x})=\langle c_{\sigma}^{*}, \mathcal{K}(\cdot,\tilde{x})\rangle_{\mathcal{H}_{\mathcal{K}}}$
with \Cref{lem:reg-reproducing} applied to $g=c_{\sigma}^{*}$,
\begin{align*}
  c_{\sigma}^{*}(\tilde{x}) - \hat{c}_{\sigma}(\tilde{x})
   = \big\langle c_{\sigma}^{*},\, \mathcal{K}(\cdot,\tilde{x})\big\rangle_{\mathcal{H}_{\mathcal{K}}}
     - \big\langle c_{\sigma}^{*},\, S^\lambda_{\mathcal{K}(\cdot,\tilde{x})}\big\rangle_{\mathcal{H}_{\mathcal{K}}}
   = \big\langle c_{\sigma}^{*},\; \mathcal{K}(\cdot,\tilde{x}) - S^\lambda_{\mathcal{K}(\cdot,\tilde{x})}\big\rangle_{\mathcal{H}_{\mathcal{K}}}.
\end{align*}
By the Cauchy--Schwarz inequality it follows that
\begin{align*}\label{eq:cs-bound}
  \big|c_{\sigma}^{*}(\tilde{x}) - \hat{c}_{\sigma}(\tilde{x})\big|
   \;\le\; \|c_{\sigma}^{*}\|_{\mathcal{H}_{\mathcal{K}}}\cdot
     \underbrace{\big\|\mathcal{K}(\cdot,\tilde{x}) - S^\lambda_{\mathcal{K}(\cdot,\tilde{x})}\big\|_{\mathcal{H}_{\mathcal{K}}}}
       _{=:\,P_{\mathcal{K},\lambda}(\tilde{x},I)} ,
\end{align*}
the \emph{regularized power function}, whose closed form we compute next.

\subsubsection*{Closed form for the regularized power function}\label{app:subsubsec:reconstr_err_closedform_reg_power}

Expanding $P_{\mathcal{K},\lambda}(\tilde{x},I)^2$ yields three summands, exactly as in \Cref{app:subsubsec:reconstr_err_closedform_power}:
\[
  P_{\mathcal{K},\lambda}(\tilde{x},I)^2
   = \underbrace{\|\mathcal{K}(\cdot,\tilde{x})\|_{\mathcal{H}_{\mathcal{K}}}^2}_{(A)}
     - 2\underbrace{\big\langle \mathcal{K}(\cdot,\tilde{x}),\,S^\lambda_{\mathcal{K}(\cdot,\tilde{x})}\big\rangle_{\mathcal{H}_{\mathcal{K}}}}_{(B)}
     + \underbrace{\|S^\lambda_{\mathcal{K}(\cdot,\tilde{x})}\|_{\mathcal{H}_{\mathcal{K}}}^2}_{(C)} .
\]
The three summands reformulate as
\begin{align*}
  (A)\;&=\; \langle \mathcal{K}(\cdot,\tilde{x}),\mathcal{K}(\cdot,\tilde{x})\rangle_{\mathcal{H}_{\mathcal{K}}}
        \;=\; \mathcal{K}(\tilde{x},\tilde{x}); \\
  (B)\;&=\; S^\lambda_{\mathcal{K}(\cdot,\tilde{x})}(\tilde{x})
        \;=\; \trans{\mathbf{k}(\tilde{x})}(\mathbf{K}+\lambda\mathbf{I}_p)^{-1}\mathbf{k}(\tilde{x}); \\
  (C)\;&=\; \trans{\boldsymbol{\beta}}\mathbf{K}\boldsymbol{\beta}
        \;=\; \trans{\mathbf{k}(\tilde{x})}(\mathbf{K}+\lambda\mathbf{I}_p)^{-1}\mathbf{K}(\mathbf{K}+\lambda\mathbf{I}_p)^{-1}\mathbf{k}(\tilde{x}) \\
       &=\; \trans{\mathbf{k}(\tilde{x})}(\mathbf{K}+\lambda\mathbf{I}_p)^{-1}\mathbf{k}(\tilde{x})
            \;-\; \lambda\,\trans{\mathbf{k}(\tilde{x})}(\mathbf{K}+\lambda\mathbf{I}_p)^{-2}\mathbf{k}(\tilde{x}),
\end{align*}
where $(C)$ uses the RKHS inner-product formula
$\langle \sum_i\alpha_i \mathcal{K}(\cdot,x_i), \sum_j\gamma_j \mathcal{K}(\cdot,x_j)\rangle_{\mathcal{H}_{\mathcal{K}}}
 = \trans{\boldsymbol{\alpha}}\mathbf{K}\boldsymbol{\gamma}$, followed by the identity
$(\mathbf{K}+\lambda\mathbf{I}_p)^{-1}\mathbf{K}(\mathbf{K}+\lambda\mathbf{I}_p)^{-1}
 = (\mathbf{K}+\lambda\mathbf{I}_p)^{-1} - \lambda(\mathbf{K}+\lambda\mathbf{I}_p)^{-2}$ (this can be seen by setting $\mathbf{K} = (\mathbf{K}+\lambda\mathbf{I}_p) - \lambda\mathbf{I}_p$).
In contrast to the unregularized case, where
$(C)=\trans{\mathbf{k}(\tilde{x})}\mathbf{K}^{-1}\mathbf{k}(\tilde{x})=(B)$ and the expansion collapses to two
terms, regularization breaks the cancellation and leaves the extra summand. Combining
$(A)-2(B)+(C)$,
\begin{align*}\label{eq:reg-power}
  \;P_{\mathcal{K},\lambda}(\tilde{x},I)
    = \Big( \mathcal{K}(\tilde{x},\tilde{x})
      - \trans{\mathbf{k}(\tilde{x})}(\mathbf{K}+\lambda\mathbf{I}_p)^{-1}\mathbf{k}(\tilde{x})
      - \lambda\,\trans{\mathbf{k}(\tilde{x})}(\mathbf{K}+\lambda\mathbf{I}_p)^{-2}\mathbf{k}(\tilde{x})
      \Big)^{1/2}.
\end{align*}
In conclusion, in the well-specified regime $\mathbf{y}_{\sigma}=c_{\sigma}^{*}$ the total error admits the single
closed-form bound
\begin{align*}
  \big|c_{\sigma}^{*}(\tilde{x}) - \hat{c}_{\sigma}(\tilde{x})\big| \;\le\; \|c_{\sigma}^{*}\|_{\mathcal{H}_{\mathcal{K}}}\,P_{\mathcal{K},\lambda}(\tilde{x},I).
\end{align*}

\subsection{Full Matrix Bound}\label{app:subsubsec:reconstr_err_full_matrix_bound}

Putting everything together for all functions $\sigma \in [s]$ and $\tilde{x} \in \tilde{I}$, we obtain the full matrix bound with the regularized power function
\[
    \| \hat{\mathbf{C}}_{\tilde{I}} - \textbf{C}^{*}_{\tilde{I}}\|^2
    = \sum_{\sigma=1}^{s}\sum_{\tilde{x}\in\tilde{I}}
      |c_{\sigma}^{*}(\tilde{x}) - \hat{c}_\sigma(\tilde{x})|^2
    \;\leq\;
    |\tilde{I}|\cdot P_{\mathcal{K},\lambda}^{\max}(\tilde{I},I)^2\cdot \|\textbf{C}^{*}\|_{\mathcal{H}_\mathcal{K}}^2,
\]
where
\[
    P_{\mathcal{K},\lambda}^{\max}(\tilde{I},I) := \max_{\tilde{x}\in\tilde{I}} P_{\mathcal{K},\lambda}(\tilde{x},I),
    \qquad
    \|\textbf{C}^{*}\|_{\mathcal{H}_\mathcal{K}}^2 := \sum_{\sigma=1}^{s}\|c_{\sigma}^{*}\|_{\mathcal{H}_{\mathcal{K}}}^2.
\]

\subsection{Full Tensor Bound}\label{app:subsubsec:reconstr_err_full_tensor_bound}

By multilinearity of the mode products, the estimated and true transfer-domain tensors share the core  $\mathbfcal{G}$ and the discrete factors $\textbf{A}, \textbf{B}$, so that 
\begin{align*}
    \widehat{\mathbfcal{T}}_{\tilde I}-\mathbfcal{T}^{*}_{\tilde I}
    &= (\mathbfcal{G} \bullet_1 \textbf{A} \bullet_2 \textbf{B} \bullet_3 \hat{\mathbf{C}}_{\tilde I}) - (\mathbfcal{G} \bullet_1 \textbf{A} \bullet_2 \textbf{B} \bullet_3 \mathbf{C}^{*}_{\tilde I}),
\end{align*}
and by mode-$3$ unfolding it follows that
\begin{align*}
    (\hat{\mathbf{C}}_{\tilde I} - \mathbf{C}^{*}_{\tilde I}) \textbf{G}_{(3)} \trans{(\textbf{B} \otimes \textbf{A})}
    &= \Delta\textbf{C} \textbf{G}_{(3)} \trans{(\textbf{B} \otimes \textbf{A})}
\end{align*}
for $\Delta\textbf{C} = \hat{\mathbf{C}}_{\tilde I} - \mathbf{C}^{*}_{\tilde I} \in\mathbb{R}^{|\tilde I|\times s}$.
Since the Frobenius norm of a tensor equals that of any unfolding and, by orthonormality of $\textbf{A}$ and $\textbf{B}$ (Assumption~\ref{ass:A_B_orth}) it follows 
\begin{align*}
|\!| \widehat{\mathbfcal{T}}_{\tilde I}-\mathbfcal{T}^{*}_{\tilde I} |\!|^{2}
= \operatorname{tr} \bigl(\Delta\textbf{C}\mathbf{G}_{(3)}\mathbf{G}_{(3)}^{\mathsf T}\Delta\textbf{C}^{\mathsf T}\bigr)
= \operatorname{tr}\bigl(\Delta\textbf{C}\textbf{M}\Delta\textbf{C}^{\mathsf T}\bigr).
\end{align*}
Diagonalizing $\textbf{M}=\textbf{Q}\Lambda\textbf{Q}^{\mathsf T}$ with $\Lambda=\operatorname{diag}(\mu_1,\dots,\mu_s)$ and $\textbf{Q}^{\mathsf T}\textbf{Q}=\textbf{I}_s$ as in Assumption~\ref{ass:core_unf_full_rank}, and setting $\Delta\textbf{C}\textbf{Q}=\hat{\textbf{C}}'_{\tilde I} - \textbf{C}'^{*}_{\tilde I}$, we obtain the decoupled identity
\begin{align*}
\bigl\|\widehat{\mathbfcal{T}}_{\tilde I}-\mathbfcal{T}^{*}_{\tilde I}\bigr\|^{2}
=\sum_{\sigma=1}^{s}\mu_\sigma\,\bigl\|\Delta\textbf{C}\,q_\sigma\bigr\|_{2}^{2}.
\end{align*}
Each column $\Delta\textbf{C}\,q_\sigma$ is the rotated continuous-factor error of a single
kernel ridge regression exactly as in the decoupling of Subsection~\ref{app:subsubsec:reconstr_err_decoupling}. With the matrix-wise bound established above and by bounding $\mu_\sigma\le\mu_s$, we obtain 
\begin{align*}
    |\!| \widehat{\mathbfcal{T}}_{\tilde I} - \mathbfcal{T}^{*}_{\tilde I} |\!|^2 \le |\tilde{I}| \mu_{s} P_{\mathcal{K},\lambda}^{\max}(\tilde{I},I)^2 \|\textbf{C}^{*}\|_{\mathcal{H}_\mathcal{K}}^2.
\end{align*}

\section{Experimental results: Additional figures}\label{app:subsec:experiments_add_info}

\subsection{Semi-synthetic data}\label{app:subsubsec:exp_synthetic_results}

\begin{figure}[!htb]
\centering
\includegraphics[width=\linewidth]{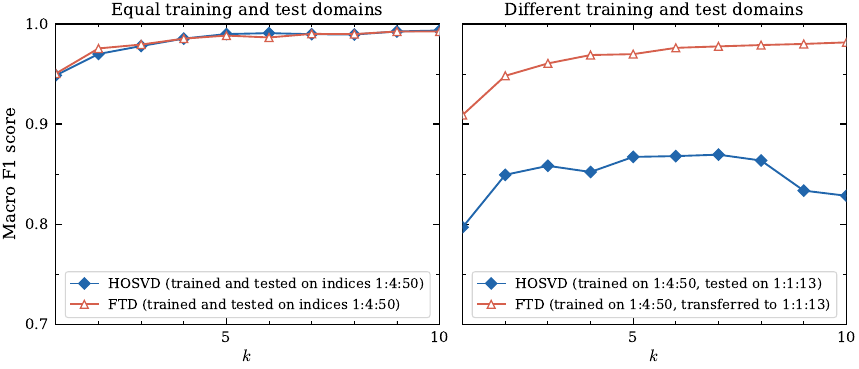}
\caption{Macro F1 scores of HOSVD- and FTD-driven classification for equal (l.h.s.) and different (r.h.s.) training and test domains for the semi-synthetic digit data. As with the classification accuracy, the FTD and HOSVD perform similarly when training and test use identical sampling points. When the sampling points differ, however, the adaptive subspace modeling allows the FTD to adapt accordingly.}
\label{app:fig:ex1_macrof1scores_lambda_1_d_4}
\end{figure}

\subsection{Multivariate time-series data}\label{app:subsubsec:exp_ts_results}

\begin{figure}[!htb]
\centering
\includegraphics[width=\linewidth]{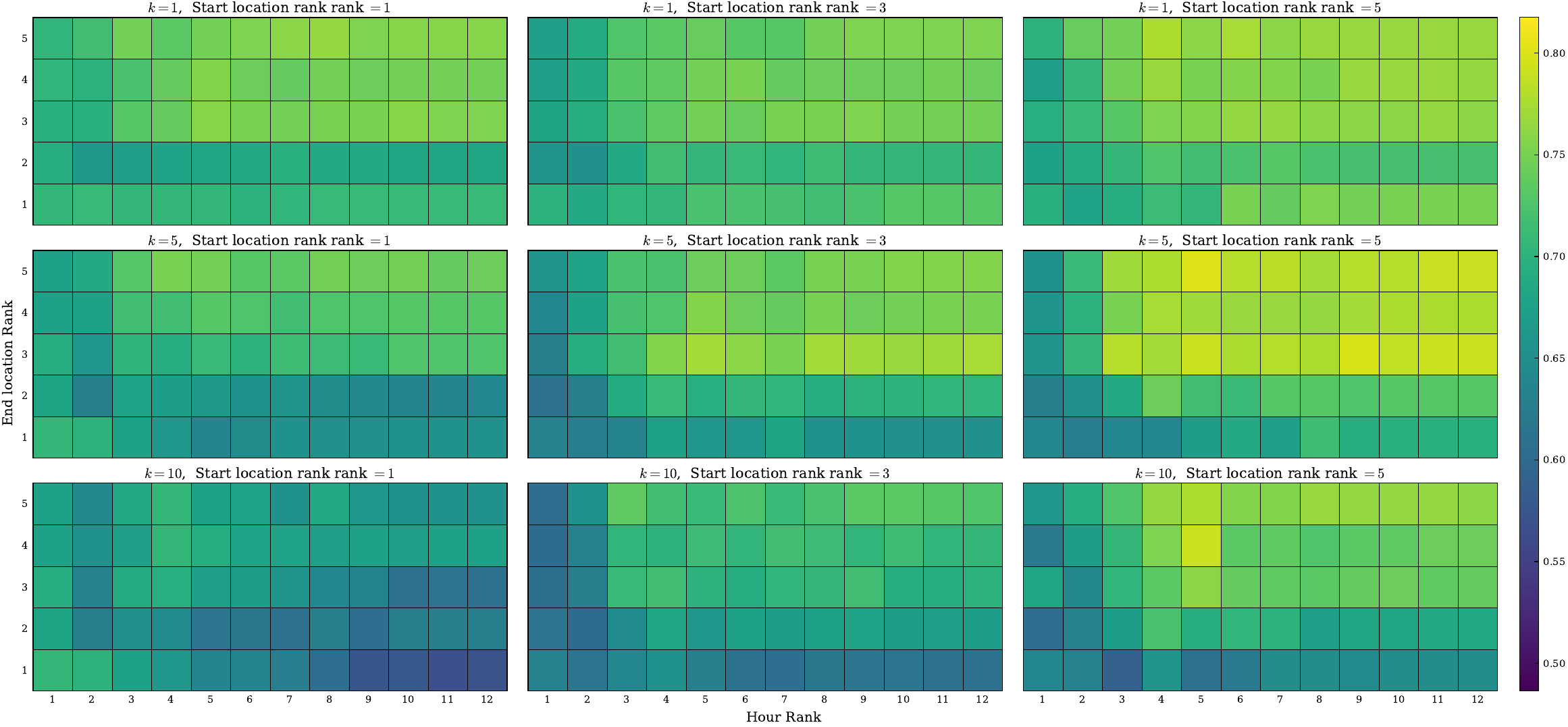}
\caption{Five-fold cross-validated HOSVD classification accuracy on the NYC taxi-trip dataset across multilinear truncation ranks. Within each panel, accuracy is shown as a function of the hour rank and the drop-off-location rank . Panels vary over the truncation rank $k\in\{5,10,15\}$ down the rows and over the pickup-location rank $\in\{1,3,5\}$ across the columns.}
\label{app:fig:ex2_ny_taxi_data_hosvd_cross_validation_k_start_loc}
\end{figure}

\begin{figure}[!htb]
\centering
\includegraphics[width=\linewidth]{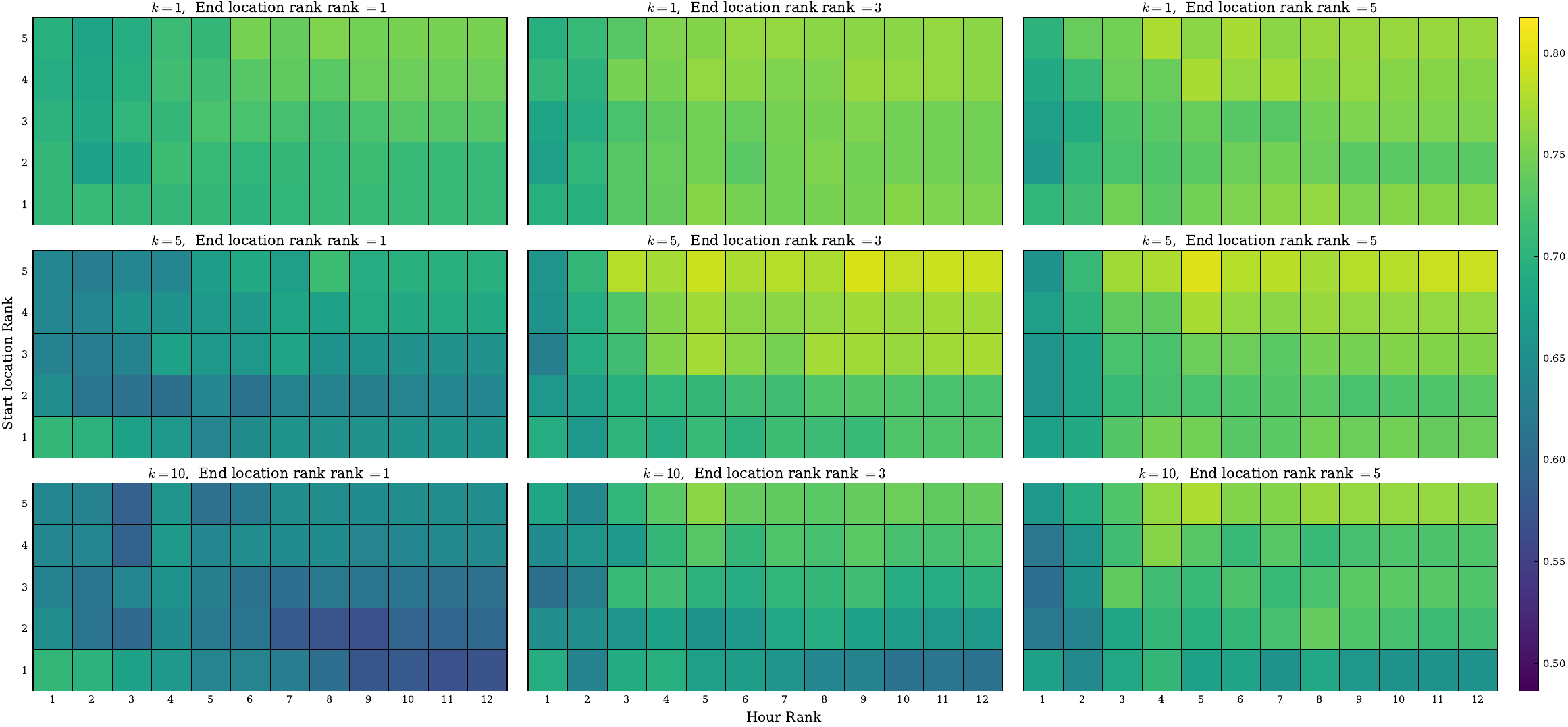}
\caption{Five-fold cross-validated HOSVD classification accuracy on the NYC taxi-trip dataset across multilinear truncation ranks. Within each panel, accuracy is shown as a function of the hour rank and the pickup-location rank . Panels vary over the truncation rank $k\in\{5,10,15\}$ down the rows and over the drop-off-location rank $\in\{1,3,5\}$ across the columns.}
\label{app:fig:ex2_ny_taxi_data_hosvd_cross_validation_k_end_loc}
\end{figure}

\begin{figure}[!htb]
\centering
\includegraphics[width=\linewidth]{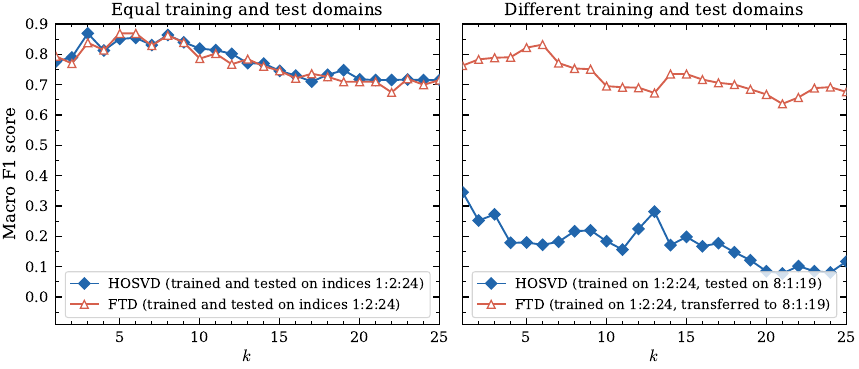}
\caption{Macro F1 scores of HOSVD- and FTD-driven classification for equal (l.h.s.) and different (r.h.s.) training and test domains for the NYC taxi trip data. The Macro F1 scores confirm the discriminative power observed in the domain‑transfer setting, consistent with the findings from the classification accuracy. When the training and test sampling domains coincide, both the FTD and the HOSVD perform similarly. In contrast, under subspace transfer, the HOSVD scores drop to approximately $0.3$, whereas the FTD maintains a stable Macro F1 score of around $0.7$ as before.}
\label{app:fig:ex2_macrof1scores_lambda_1_d_1}
\end{figure}

\subsection{Hyperspectral image data}\label{app:subsubsec:exp_hsi_results}

\begin{figure}[!htb]
\centering
\includegraphics[width=\linewidth]{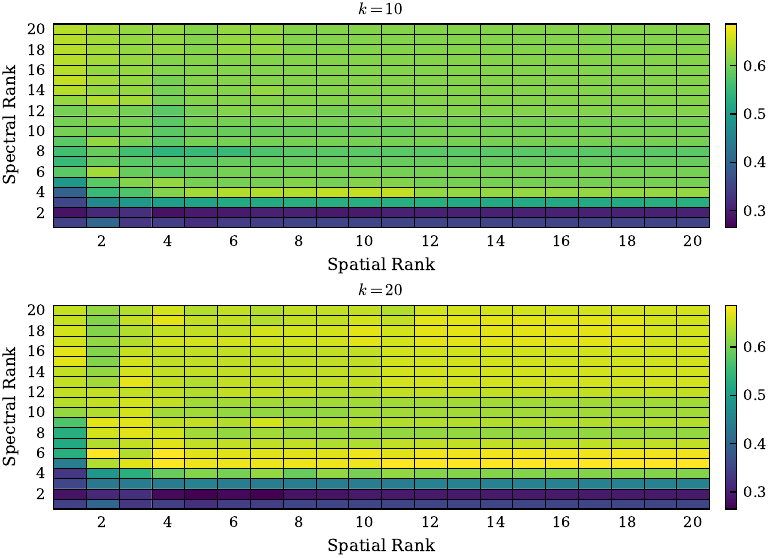}
\caption{HOSVD classification accuracy on validation set of the mango HSI data for different multi-linear truncations, varying the rank $k\in\{10, 20\}$ in the sample mode together with the truncations in spatial and spectral modes.}
\label{app:fig:ex3_hsi_mango_data_hosvd_cross_validation}
\end{figure}

\begin{figure}[!htb]
\centering
\includegraphics[width=\linewidth]{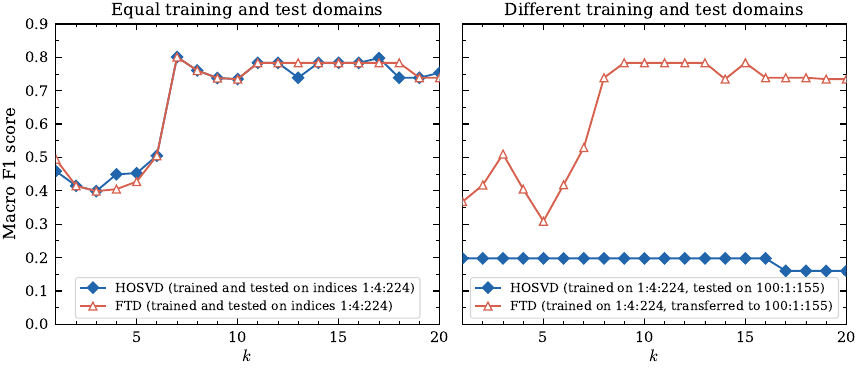}
\caption{Macro F1 scores of HOSVD- and FTD-driven classification for equal (l.h.s.) and different (r.h.s.) training and test domains for the mango HSI data. FTD and HOSVD achieve comparable performance when training and testing occur in the same domain. Under subspace transfer, however, the scores split: FTD still indicates discriminative classification power, while HOSVD fails to demonstrate any meaningful classification ability.}
\label{app:fig:ex3_macrof1scores_lambda_1_d_1}
\end{figure}

\end{document}